\documentclass[10pt,twocolumn,letterpaper]{article}

\usepackage{cvpr}
\usepackage{times}
\usepackage{epsfig}
\usepackage{graphicx}
\usepackage{amsmath}
\usepackage{amssymb}
\usepackage{array}
\usepackage{subfigure}
\newcolumntype{L}[1]{>{\raggedright\let\newline\\\arraybackslash\hspace{0pt}}m{#1}}
\newcolumntype{C}[1]{>{\centering\let\newline\\\arraybackslash\hspace{0pt}}m{#1}}
\newcolumntype{R}[1]{>{\raggedleft\let\newline\\\arraybackslash\hspace{0pt}}m{#1}}

\definecolor{darkgreen}{RGB}{10,150,10}

\definecolor{ningcolor}{rgb}{0,0,0}
\newcommand{\ning}[1]{\textcolor{ningcolor}{#1}}

\newcommand{\sectnum    } [1] {\ref{#1}}

\newcommand{\sect       } [1] {Section~\sectnum{#1}}

\newcommand{\fignum     } [1] {\ref{#1}}
\newcommand{\fig        } [1] {Figure~\fignum{#1}}

\newcommand{\vshrink}[0]{\vspace{-0.1cm}}

\usepackage[pagebackref=true,breaklinks=true,letterpaper=true,colorlinks,bookmarks=false]{hyperref}

\cvprfinalcopy 

\def\cvprPaperID{5947} 

\ifcvprfinal\fi
\begin{document}

\title{Texture~Mixer:~A~Network for Controllable Synthesis and~Interpolation~of~Texture}

\author{Ning Yu\hspace{0.01in}\textsuperscript{1,2,4}\hspace{3ex}
Connelly Barnes\hspace{0.01in}\textsuperscript{3,4}\hspace{3ex}
Eli Shechtman\hspace{0.01in}\textsuperscript{3}\hspace{3ex}
Sohrab Amirghodsi\hspace{0.01in}\textsuperscript{3}\hspace{3ex}
Michal Luk\'{a}\v{c}\hspace{0.01in}\textsuperscript{3}\\
\hspace{0.2in}\textsuperscript{1}\hspace{0.01in}University of Maryland
\hspace{3ex} \textsuperscript{2}\hspace{0.01in}Max Planck Institute for Informatics\\
\hspace{3ex} \textsuperscript{3}\hspace{0.01in}Adobe Research
\hspace{3ex} \textsuperscript{4}\hspace{0.01in}University of Virginia\\
{\tt\small ningyu@mpi-inf.mpg.de}
\hspace{3ex}{\tt\small connelly@cs.virginia.edu}
\hspace{3ex}{\tt\small \{elishe, tamirgho, lukac\}@adobe.com}\\
}

\maketitle
\thispagestyle{empty}

\begin{abstract}
This paper addresses the problem of interpolating visual textures. We formulate this problem by requiring (1) by-example controllability and (2) realistic and smooth interpolation among an arbitrary number of texture samples. To solve it we propose a neural network trained simultaneously on a reconstruction task and a generation task, which can project texture examples onto a latent space where they can be linearly interpolated and projected back onto the image domain, thus ensuring both intuitive control and realistic results. We show our method outperforms a number of baselines according to a comprehensive suite of metrics as well as a user study. We further show several applications based on our technique, which include texture brush, texture dissolve, and animal hybridization~\footnote{Demos, videos, code, data, models, and supplemental material are available at \href{https://github.com/ningyu1991/TextureMixer.git}{GitHub}.}.
\end{abstract}

\vshrink{}
\section{Introduction}


\begin{figure}
\centering
\includegraphics[width=3.25in]{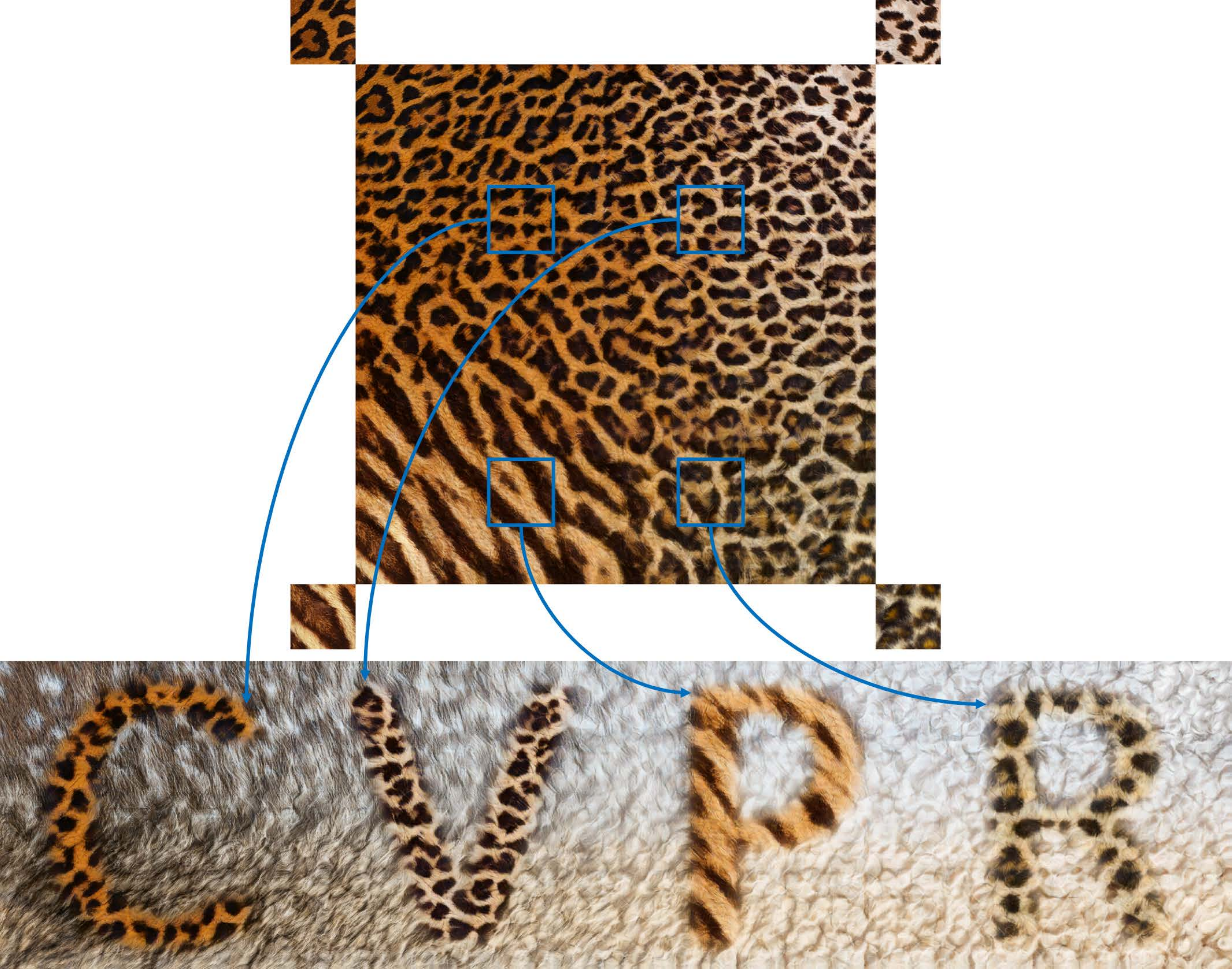}
\caption{Texture interpolation and texture painting using our network on the \emph{animal texture} dataset. The top part shows a $1024 \times 1024$ palette created by interpolating four source textures at the corners outside the palette. The bottom part shows a $512 \times 2048$ painting of letters with different textures sampled from the palette. The letters are interpolated by our method with the background, also generated by our interpolation.}
\label{fig:motivation}
\vspace{-12pt}
\end{figure}


Many materials exhibit variation in local appearance, as well as complex transitions between different materials. Editing materials in an image, however, can be highly challenging due to the rich, spatially-varying material combinations as we see in the natural world. One general research challenge then is to attempt to enable these kinds of edits. In particular, in this paper, we focus on textures. We define ``texture" as being an image-space representation of a statistically homogeneous material, captured from a top-down view. We further focus on allowing a user to both be able to accurately control the placement of textures, as well as create plausible transitions between them.

Because of the complex appearance of textures, creating transitions by interpolating between them on the pixel domain is difficult. Doing so na\"{\i}vely results in unpleasant artifacts such as ghosting, visible seams, and obvious repetitions. Researchers in texture synthesis have therefore developed sophisticated algorithms to address this problem. These may be divided to two families: non-parametric methods such as patch-based synthesis (e.g.~\cite{efros1999texture,efros2001image,barnes2009patchmatch}) and parametric methods (e.g.~\cite{heeger1995pyramid, portilla2000parametric}), including neural network synthesis approaches (e.g.~\cite{gatys2015texture,ulyanov2016texture, johnson2016perceptual, li2016precomputed, li2017diversified}). Previously, researchers used sophisticated patch-based interpolation methods~\cite{darabi2012image,diamanti2015synthesis} with carefully crafted objective functions. However, such approaches are extremely slow. Moreover, due to the hand-crafted nature of their objectives, they cannot learn from a large variety of textures in the natural world, and as we show in our comparisons are often brittle and frequently result in less pleasing transitions. Further, we are not aware of any existing feedforward neural network approaches that offer both fine-grained controllable synthesis and interpolation between multiple textures. \ning{User-controllable texture interpolation is substantially more challenging than ordinary texture synthesis, because it needs to incorporate adherence to user-provided boundary conditions and a smooth transition for the interpolated texture.} 

In our paper, we develop a neural network approach that we call ``Texture Mixer," which allows for both user control and interpolation of texture. We define the \emph{interpolation} of texture as a broad term, encompassing any combination of: (1) Either gradual or rapid \emph{spatial transitions} between two or more different textures, as shown in the palette, the letters, and the background in \fig{fig:motivation}, and (2) \emph{Texture dissolve}, where we can imagine putting two textures in different layers, and cross-dissolving them according to a user-controlled transparency, as we show in our video. \ning{Previous neural methods can create interpolations similar to our dissolves by changing the latent variable~\cite{huang2017arbitrary,karras2017progressive,li2017diversified,li2017universal,chen2017stylebank}. Thus, in this paper we focus primarily on high-quality spatial interpolation: this requires textures to coexist in the same image plane without visible seams or spatial repetitions, which is more difficult to achieve.} Our feedforward network is trained on a large dataset of textures and runs at interactive rates.

Our approach addresses the difficulty of interpolating between textures on the image domain by projecting these textures onto a latent domain where they may be linearly interpolated, and then decoding them back into the image domain to obtain the desired result. In order to satisfy the two goals of \emph{controllability} and \emph{visual realism}, we train our network simultaneously for both tasks. A \emph{reconstruction task} ensures that when a texture is passed through an encoder and then a decoder (an autoencoder), the result will be similar to the input. This allows the user to specify texture at any given point of the output by example. 
An \emph{interpolation task} uses a discriminator to ensure that linear interpolations of latent tensors also decode into plausible textures, so that the regions of the output not directly specified by the user are realistic and artifact-free. For this task, we can view our network as a conditional Generative Adversarial Network (GAN). In effect, we thus train an autoencoder and a conditional GAN at the same time, using shared weights and a shared latent space. 

To perform the interpolation task, we take texture samples that user specifies, and project them into latent space using a learned encoder. Given these latent tensors, our network then uses three intuitive latent-space operations: tiling, interpolation, and shuffling. The tiling operation extends a texture spatially to any arbitrary size. The interpolation operation uses weighted combinations of two or more textures in latent domain. The shuffling operation swaps adjacent small squares within the latent tensor to reduce repetitions. These new latent tensors are then decoded to obtain the interpolated result.

Our main contributions are: (1) a novel interactive technique that allows both user control and interpolation of texture; (2) several practical and creative applications based on our technique; (3) a new suite of metrics that evaluate user controllability, interpolation smoothness, and interpolation realism; and (4) the state-of-the-art performance superior to previous work both based on these metrics, and based on a user study if we consider them holistically.

\begin{figure*}[!ht]
\centering
\includegraphics[width=0.95\linewidth]{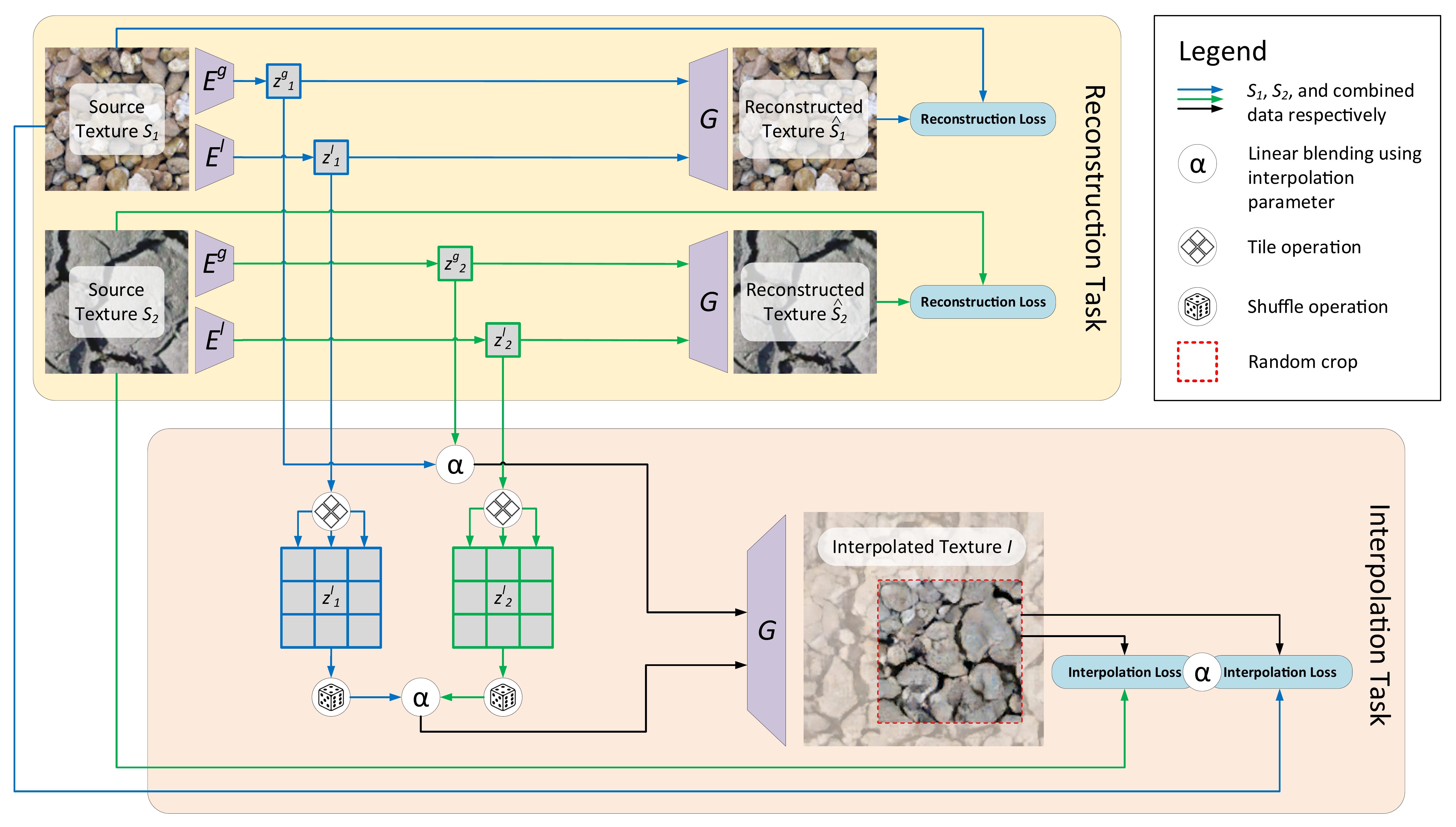}
\caption{A diagram of our method. Background color highlights each of the tasks. Trapezoids represent trainable components that share weights if names match. Rounded rectangles represent the losses. Arrows and circles represent operations on tensor data.
}
\label{fig:pipeline_diagram}
\vspace{-12pt}
\end{figure*}

\vshrink{}
\section{Related Work}
The problem of user-controllable texture interpolation has so far been under-explored. It is however closely related to several other problems, most significantly texture synthesis, inpainting, and stylization.

Texture synthesis algorithms can be divided into two families. The first one is parametric, with a generative texture model. These algorithms include older, non-neural methods~\cite{heeger1995pyramid, portilla2000parametric}, and also more recent deep learning-based methods that are based on optimization~\cite{gatys2015texture, gatys2016image, risser2017stable, sendik2017deep} or trained feedforward models~\cite{ulyanov2016texture, johnson2016perceptual, li2016precomputed, li2017diversified}. Where the underlying model allows spatially varying weights for combination, it may be used to cross-dissolve textures. \ning{However, we are not aware of any existing texture synthesis techniques in this family that enables spatial transition between different textures.}

The second family of texture synthesis algorithms is non-parametric, in which the algorithm produces output that is optimized to be as close as possible to the input under some appearance measure~\cite{efros1999texture, wei2000fast, efros2001image, kwatra2003graphcut, kwatra2005texture, matusik2005texture, lefebvre2006appearance, wexler2007space, barnes2009patchmatch, darabi2012image, kaspar2015self}. These can be formulated to accept two different inputs and spatially vary which is being compared to, facilitating interpolation~\cite{darabi2012image, diamanti2015synthesis}. \ning{As we mentioned before, such approaches are slow, and due to the hand-crafted nature of their objectives, they tend to be brittle.}

Recently, generative adversarial networks (GANs)~\cite{goodfellow2014generative, salimans2016improved,arjovsky2017wasserstein, gulrajani2017improved} have shown improved realism in image synthesis and translation tasks~\cite{isola2017image, CycleGAN2017, zhu2017toward}. GANs have also been used directly for texture synthesis~\cite{li2016precomputed, jetchev2016texture,zhou2018nonstationary}, however, they were limited to a single texture they were trained on. A recent approach dubbed PSGAN~\cite{pmlr-v70-bergmann17a} learns to synthesize a collection of textures present in a single photograph, making it more general and applicable to texture interpolation; it is not, however, designed for our problem as it cannot interpolate existing images. \ning{We show comparisons with PSGAN and it cannot reconstruct many input textures, even after running a sophisticated optimization or jointly associating PSGAN with an encoder. Moreover, PSGAN can suffer from mode collapse.}

Texture synthesis and image inpainting algorithms are often closely related. A good hole filling algorithm needs to be able to produce some sort of transition between textures on opposite ends of the hole, and so may be used in a texture interpolation task. A few recent deep learning-based methods showed promising results~\cite{yang2017high, yu2018generative, liu2018image, yu2018free}.

Finally, some neural stylization approaches~\cite{gatys2016image,li2016precomputed, huang2017arbitrary,li2017universal} based on separating images into content and style components have shown that, by stylizing a noise content image, they can effectively synthesize texture~\cite{gatys2015texture}. By spatially varying the style component, texture interpolation may thus be achieved.

\vshrink{}
\section{Our network: Texture Mixer}

In this section, we explain how our network works. We first explain in \sect{sec:training} how our method is trained. We then show how our training losses are set up in \sect{sec:training_loss}. Finally, we explain in \sect{sec:testing} how our method can be either tested or used by an end user.

\subsection{Training setup}
\label{sec:training}
We aim to train our network simultaneously for two tasks:  \emph{reconstruction} and \emph{interpolation}. The \emph{reconstruction task} ensures that every input texture after being encoded and then decoded results in a similar texture. Meanwhile, the \emph{interpolation task} ensures that interpolations of latent tensors are also decoded into plausible textures. 

Our method can be viewed as a way of training a network containing both encoders and a generator, such that the generator is effectively a portion of a GAN. The network accepts a source texture $S$ as input. A \emph{global encoder} $E^g(S)$ encodes $S$ into a latent vector $z^g$, which can also be viewed as a latent tensor with spatial size $1\times 1$. A \emph{local encoder} $E^l(S)$  encodes the source texture into a latent tensor $z^l$, which has a spatial size that is a factor $m$ smaller than the size of the input texture: we use $m = 4$. The generator $G(z^l,z^g)$ \ning{concatenates $z^l$ and $z^g$, and} can decode these latent tensors back into a texture patch, so that ideally $G(E^l(S),E^g(S))=S$, which encompasses the reconstruction task. Our generator is fully convolutional, so that it can generate output textures of arbitrary size: the output texture size is directly proportional to the size of the local tensor $z^l$. A discriminator $D^{\text{rec}}$ is part of the reconstruction loss. An identical but separately trained discriminator $D^{\text{itp}}$ evaluates the realism of interpolation.

Note that in practice, our generator network is implemented as taking a global tensor as input, which has the same spatial size as the local tensor. This is because, for some applications of texture interpolation, $z^g$ can actually vary spatially. Thus, when we refer to $G$ taking a global latent vector $z^g$ with spatial size $1\times 1$ as input, what we mean is that this $z^g$ vector is first repeated spatially to match the size of $z^l$, and the generator is run on the result. 

We show the full training setup in \fig{fig:pipeline_diagram}. We will also explain \ning{our setup} in terms of formulas here. As is shown in the upper-left of \fig{fig:pipeline_diagram}, the network is given two real source texture images $S_1$ and $S_2$ from the real texture dataset $\mathcal{S}$. Each local encoder $E^l$ encodes $S_i$ ($i \in \{1, 2\}$) to a local latent tensor $z^l_i = E^l(S_i)$. Meanwhile, each global encoder $E^g$ encodes $S_i$ to a global latent vector $z^g_i$, denoted as $z^g_i = E^g(S_i)$. These latent variables are shown in green and blue boxes in the upper-left of \fig{fig:pipeline_diagram}.

For the {\em reconstruction task}, we then evaluate the reconstructed texture image $\hat{S_i} = G\big(z^l_i, z^g_i)$. These are shown in the upper center of \fig{fig:pipeline_diagram}. For each reconstructed image $\hat{S_i}$, we then impose a weighted sum of three losses against the original texture $S_i$. We describe these losses in more detail later in \sect{sec:training_loss}.

For the {\em interpolation task}, we pose the process of multiple texture interpolation as a problem of simultaneously (1) synthesizing a larger texture, and (2) interpolating between two different textures. In this manner, the network learns to perform well for both single and multiple texture synthesis. For single texture synthesis, we enlarge the generated images by a factor of $3 \times 3$. We do this by tiling $z^l_i$ spatially by a factor of $3 \times 3$. We denote this tiling by $T(z^l_i)$, and indicate tiling by a tile icon in the lower-left of \fig{fig:pipeline_diagram}. We chose the factor 3 because this is the smallest integer that can synthesize transitions over the four edges of $z^l_i$. Such a small tiling factor minimizes computational cost. The tiling operation can be beneficial for regular textures. However, in semiregular or stochastic textures, the tiling introduces two artifacts: undesired spatial repetitions, and undesired seams on borders between tiles. 

We reduce these artifacts by applying a random shuffling to the tiled latent tensors $T(z^l_i)$. In \fig{fig:pipeline_diagram}, this shuffling operation is indicated by a dice icon. Random shuffling in the latent space not only results in more varied decoded image appearance and thus reduces visual repetition, but also softens seams by spatially swapping “pixels” in the latent space across the border of two $z^l_i$ tensors.

We implement the random shuffling by row and column swapping over several scales from coarse to fine. For this coarse to fine process, we use scales that are powers of two: $s_i = 2^i$ for $i=0, 2, \ldots, n$. We set the coarsest scale $n$ to give a scale $s_n$ that is half the size of the local tensor $z^l_i$. For each scale $s_i$, we define a grid over the tiled latent tensor $T(z^l)$, where each grid cell has size $s_i \times s_i$. For each scale $s_i$, we then apply a random shuffling on cells of the grid for that scale: we denote this by $P_{i}$. This shuffling proceeds through grid rows first in top-down and then bottom-up order: each row is randomly swapped with the succeeding row with probability 0.5. Similarly, this is repeated on grid columns, with column swapping from left to right and right to left. Thus, the entire shuffling operation is:
\begin{equation}
P\big(T(z^l_i)\big) = P_{0}\circ P_{1}\circ\cdots\circ P_{n}\big(T(z^l_i)\big)
\label{permutation}
\end{equation}
We visualize this shuffling procedure in the supplementary material. We also want the synthesized texture to be able to transit smoothly between regions where there are user-specified texture constraints and regions where there are none. Thus, we override the original $z^l_i$ without shuffling at the 4 corners of the tiled latent tensor. We denote such shuffling with corner overriding as $\tilde{P}\big(T(z^l_i)\big)$.

If we apply the fully convolutional generator $G$ to a network trained using a single input texture and the above shuffling process, it will work for single texture synthesis. However, for multiple texture interpolation, we additionally apply interpolation in the latent space before calling $G$, as inspired by \cite{li2017diversified, huang2017arbitrary, pmlr-v70-bergmann17a}. We randomly sample an interpolation parameter $\alpha \sim U[0, 1]$, and then interpolate the latent tensors using $\alpha$. This is shown by the circles labeled with $\alpha$ in \fig{fig:pipeline_diagram}. We linearly blend the shuffled local tensors $\tilde{P}\big(T(z^l_1)\big)$ and $\tilde{P}(T(z^l_2))\big)$, which results in the final interpolated latent tensor $Z^l$:
\begin{equation}
Z^l = \alpha\tilde{P}\big(T(z^l_1)\big) + (1 - \alpha)\tilde{P}\big(T(z^l_2)\big)
\end{equation}
In the same way, we blend $z^g_1$ and $z^g_2$ to obtain
\begin{equation}
Z^g = \alpha z^g_1 + (1 - \alpha)z^g_2
\end{equation}
Finally, we feed the tiled and blended tensors into the generator $G$ to obtain an interpolated texture image $I = G(Z^l, Z^g)$, which is shown on the right in \fig{fig:pipeline_diagram}. From the interpolated texture, we take a random crop of the same size as the input textures. The crop is shown in the red dotted lines in \fig{fig:pipeline_diagram}. The crop is then compared using appropriately $\alpha$-weighted losses to each of the source textures. \ning{We use spatially uniform weights $\alpha$ at training time because all the real-world examples are \textit{spatially homogeneous} and we do not want our adversarial discriminator to detect our synthesized texture due to it having spatial variation. In contrast, at testing time, we use spatially varying weights.}

\begin{figure*}
\centering
\includegraphics[width=1\linewidth]{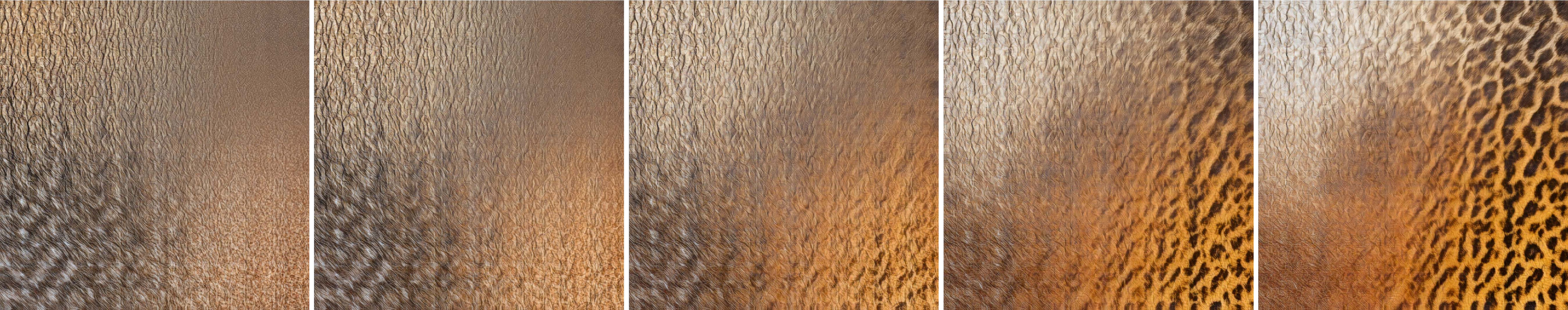}
\caption{A sequence of dissolve video frame samples with size $1024 \times 1024$ on the \emph{animal texture} dataset, where each frame is also with effect of interpolation.}
\label{fig:dissolve}
\vspace{-12pt}
\end{figure*}

\subsection{Training losses}
\label{sec:training_loss}

For the \emph{reconstruction task}, we use three losses. The first loss is a pixel-wise $L_1$ loss against each input $S_i$. The second loss is a Gram matrix loss against each input $S_i$, based on an ImageNet-pretrained VGG-19 model. We define the Gram loss $L_{\text{Gram}}$ in the same manner as Johnson~\etal~\cite{johnson2016perceptual}, and use the features $\text{relu}i\_1$ for $i=1, \ldots, 5$. The third loss is an adversarial loss $L_{\text{adv}}$ based on WGAN-GP~\cite{gulrajani2017improved}, where the reconstruction discriminator $D^{\text{rec}}$ tries to classify whether the reconstructed image is from the real source texture set or generated by the network. The losses are:
\begin{equation}
L^{\text{rec}}_{\text{pix}} = \|\hat{S_1} - S_1\|_1 + \|\hat{S_2} - S_2\|_1
\label{rec_L1}
\end{equation}\vspace{-0.5cm}
\begin{equation}
L^{\text{rec}}_{\text{Gram}} = L_{\text{Gram}}(\hat{S_1},S_1) + L_{\text{Gram}}(\hat{S_2},S_2)
\label{rec_Gram}
\end{equation}
\begin{equation}
L^{\text{rec}}_{\text{adv}} = L_{\text{adv}}(\hat{S_1}, S_1 | D^{\text{rec}}) + L_{\text{adv}}(\hat{S_2}, S_2 | D^{\text{rec}})
\label{rec_adv}
\end{equation}

The $L_{\text{adv}}$ term is defined from WGAN-GP~\cite{gulrajani2017improved} as:
\begin{equation}
L_{\text{adv}}(A, B | D) = D(A) - D(B) + GP(A, B | D)
\end{equation}
Here $A$ and $B$ are a pair of input images, $D$ is the adversarially trained discriminator, and $GP(\cdot)$ is the gradient penalty regularization term.

For the \emph{interpolation task}, we expect the large interpolated texture image to be similar to some combination of the two input textures. Specifically, if $\alpha = 1$, the interpolated image should be similar to source texture $S_1$, and if $\alpha = 0$, it should be similar to $S_2$. However, we do not require pixel-wise similarity, because that would encourage ghosting. We thus impose only a Gram matrix and an adversarial loss. We select a random crop $I_{\text{crop}}$ from the interpolated texture image. Then the Gram matrix loss for interpolation is defined as an $\alpha$-weighted loss to each source texture:
\begin{equation}
L^{\text{itp}}_{\text{Gram}} = \alpha L_{\text{Gram}}(I_{\text{crop}}, S_1) + (1 - \alpha)L_{\text{Gram}}(I_{\text{crop}}, S_2)
\label{itp_Gram}
\end{equation}

Similarly, we adversarially train the interpolation discriminator $D^{\text{itp}}$ for the interpolation task to classify whether its input image is from the real source texture set or whether it is a synthetically generated interpolation:
\begin{equation}
L^{\text{itp}}_{\text{adv}} = \alpha L_{\text{adv}}(I_{\text{crop}}, S_1 | D^{\text{itp}}) + (1 - \alpha)L_{\text{adv}}(I_{\text{crop}}, S_2 | D^{\text{itp}})
\label{itp_adv}
\end{equation}

Our final training objective is
\begin{equation}
\begin{split}
\min_{E^l, E^g, G}\;\max_{D^{\text{rec}}, D^{\text{itp}}}\;\underset{S_1, S_2 \sim \mathcal{S}}{\mathbb{E}}\;(\lambda_1 L^{\text{rec}}_{\text{pix}} + \lambda_2 L^{\text{rec}}_{\text{Gram}} + \lambda_3 L^{\text{rec}}_{\text{adv}}\\
+ \lambda_4 L^{\text{itp}}_{\text{Gram}} + \lambda_5 L^{\text{itp}}_{\text{adv}})
\end{split}
\label{objective}
\end{equation}
where $\lambda_1 = 100$, $\lambda_2 = \lambda_4 = 0.001$, and $\lambda_3 = \lambda_5 = 1$ are used to balance the order of magnitude of each loss term, which are not sensitive to dataset.

We provide details related to our training and architecture in the supplementary document, such as how we used progressive growing during training~\cite{karras2017progressive}.

\subsection{Testing and user interactions}
\label{sec:testing}

At testing time, we can use our network in several different ways: we can interpolate sparsely placed textures, brush with textures, dissolve between textures, and hybridize different animal regions in one image. \ning{Each of these applications utilizes spatially varying  interpolation weights.}

\textbf{Interpolation of sparsely placed textures.} This option is shown in the palette and background in \fig{fig:motivation}. In this scenario, one or more textures are placed down by the user in the image domain. These textures are each encoded to latent domain. 

\ning{In most cases, given input textures, our method is able to achieve inherent boundary matching and continuity. However, because of the trade-off between reconstruction and interpolation losses, there might be a {\em slight} mismatch in some cases.} To make the textures better agree at boundary conditions, we postprocess our images as follows. Suppose that the user places a source textured region as a boundary condition. We first replace the reconstructed regions with the source texture. Then, within the source texture, we use graph cuts \cite{kwatra2003graphcut} to determine an optimal seam where we can cut between the source texture and the reconstruction. Finally, we use Poisson blending \cite{perez2003poisson} to minimize the visibility of this seam.

\textbf{Texture brush.} We can allow the user to brush with texture as follows. We assume that there is a textured background region, which we have encoded to latent space. The user can select any texture to brush with, \ning{by first encoding the brush texture and then brushing in the latent space}. For example, in \fig{fig:motivation} we show an example of selecting a texture from a palette created by interpolating four sparsely created textures. We find the brush texture's latent domain tensors, and apply them using a Gaussian-weighted brush. Here full weight in the brush causes the background latent tensors to be replaced entirely, and other weights cause a proportionately decreased effect. \ning{The brush can easily be placed spatially because the latent and image domains are aligned with a resizing factor $m$ related to the  architecture.} We show more results in the supplementary material.

\textbf{Texture dissolve.} We can create a cross-dissolve effect between any two textures by encoding them both to latent domain and then blending between them using blending weights that are spatially uniform. This effect is best visualized in a video, where time controls the dissolve effect. Please see our supplementary video for such results. \fig{fig:dissolve} shows a sequence of video frame samples with gradually varying weights. 

\begin{figure*}
\centering
\includegraphics[width=1\linewidth]{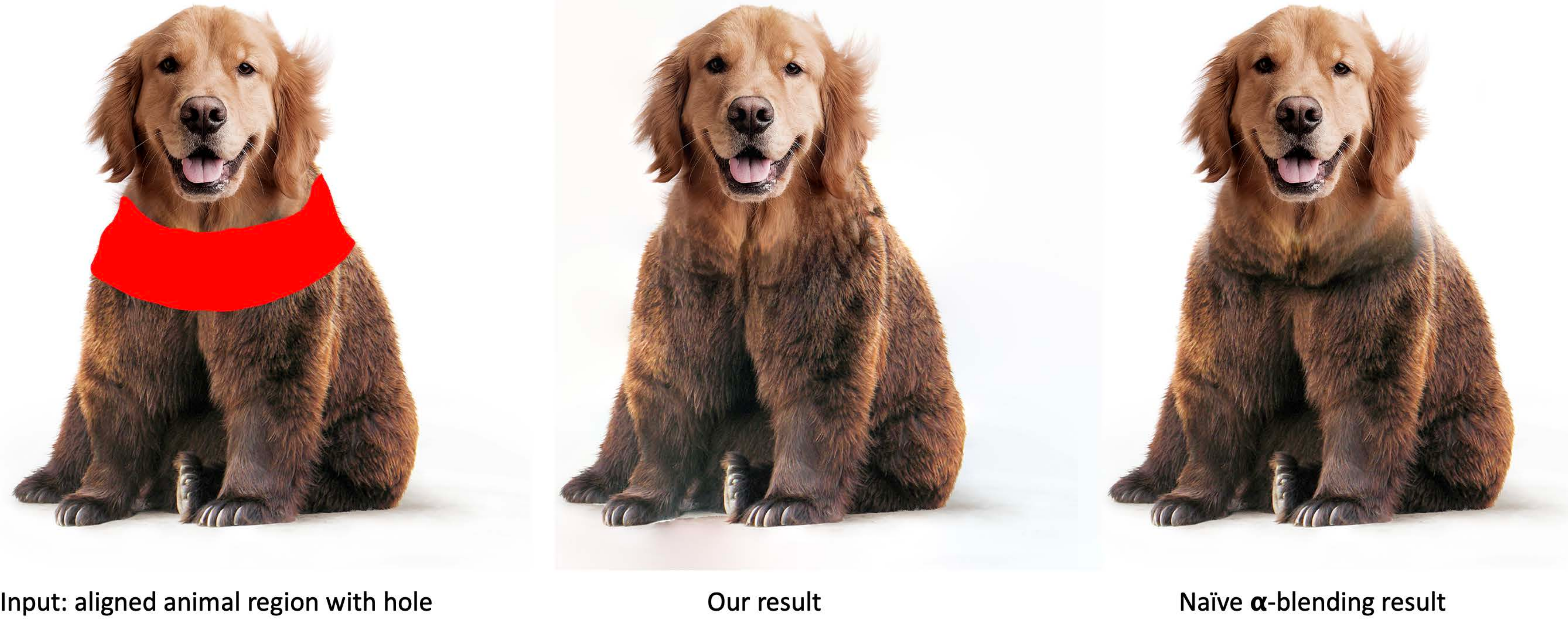}
\caption{An animal hybridization example of size $1260 \times 1260$ between a dog and a bear. Our interpolation between the two animal furs is smoother, has less ghosting, and is more realistic than that of the Na\"{\i}ve $\alpha$-blending.}
\label{fig:animal_hybridization}
\vspace{-12pt}
\end{figure*}

\ning{\textbf{Animal hybridization.} We generalize texture interpolation into a more practical and creative application - animal hybridization. \fig{fig:animal_hybridization} shows an example. Given two aligned animal regions in one image and a hole over the transition region, we can sample source texture patches adjacent to the hole and conduct spatial interpolation among those textures. We fill the hole using our interpolated texture. Finally, we use graph cuts \cite{kwatra2003graphcut} and Poisson blending \cite{perez2003poisson} to postprocess the boundaries. Technical details and more examples are shown in the supplemental material.}

\vshrink{}
\section{Experiments}
\ning{In this section, we demonstrate experimental comparisons. We first introduce our own datasets in \sect{sec:datasets}. We then present in \sect{sec:evaluation} a suite of evaluation metrics for interpolation quality. In \sect{sec:comparisons} we list and compare against several leading methods from different categories on the task of texture interpolation. In \sect{sec:user_study} we describe a user study as a holistic comparison. Finally, we conduct in \sect{sec:ablation_study} the ablation study by comparing against three simplified versions of our own method.}

\ning{We propose to learn a model per texture category rather than a universal model because: (1) there are no real-world examples that depict interpolation between distinct texture categories; (2) there is no practical reason to interpolate across categories, e.g., fur and gravel; and (3) like with other GANs, a specific model per category performs better than a universal one due to the model's capacity limit.}

\subsection{Datasets}
\label{sec:datasets}
Training to interpolate frontal-parallel stationary textures of a particular category requires a dataset with a rich set of examples to represent the intra-variability of that category. Unfortunately, most existing texture datasets such as DTD~\cite{cimpoi14describing} are intended for texture classification tasks, and do not have enough samples per category (only 120 in the case of DTD) to cover the texture appearance space with sufficient density.

Therefore, we collected two datasets of our own: (1) the \emph {earth texture} dataset contains Creative Commons images from Flickr, which we randomly split into $896$ training and $98$ testing images; (2) the \emph{animal texture} dataset contains images from Adobe Stock, randomly split into $866$ training and $95$ testing images. All textures are real-world RGB photos with arbitrary sizes larger than $512 \times 512$. Examples from both are shown in our figures throughout the paper.

We further augmented all our training and testing sets by applying: (1) color histogram matching with a random reference image in the same dataset; (2) random geometric transformations including horizontal and vertical mirroring, random in-plane rotation and downscaling (up to $\times 4$); and (3) randomly cropping a size of $128 \times 128$. In this way, we augmented $1,000$ samples for each training image and $100$ samples for each testing image. 

\subsection{Evaluation}
\label{sec:evaluation}
We will compare previous work with ours, and also do an ablation study on our own method. In order to fairly compare all methods, we use a horizontal interpolation task. Specifically, we randomly sampled two $128 \times 128$ squares from the test set. We call these the side textures. We placed them as constraints on either end of a $128 \times 1024$ canvas. We then used each method to produce the interpolation on the canvas, configuring each method to interpolate linearly where such option is available.

To the best of our knowledge, there is no standard method to quantitatively evaluate texture interpolation. We found existing generation evaluation techniques~\cite{salimans2016improved,heusel2017gans,binkowski2018demystifying,karras2017progressive} inadequate for our task. We, therefore, developed a suite of metrics that evaluate three aspects we consider crucial for our task: (1) user controllability, (2) interpolation smoothness, and (3) interpolation realism. We now discuss these.

{\bf User controllability}. For interpolation to be considered controllable, it has to closely reproduce the user's chosen texture at the user's chosen locations. In our experiment, we measure this as the reconstruction quality for the side textures. We average the LPIPS perceptual similarity measure~\cite{zhang2018perceptual} for the two side textures. We call this \emph{Side Perceptual Distance (SPD)}.

We also would like the center of the interpolation to be similar to both side textures. To measure this, we consider the Gram matrix loss~\cite{johnson2016perceptual} between the central $128 \times 128$ crop of the interpolation and the side textures. We report the sum of distances from the center crop to the two side textures, normalized by the Gram distance between the two. We call this measure the \emph{Center Gram Distance (CGD)}.

{\bf Interpolation smoothness}. Ideally, we would like the interpolation to follow the shortest path between the two side textures. To measure this, we construct two difference vectors \ning{of Gram matrix features} between the left side texture and the center crop, and between the center crop and the right side texture, and measure the cosine distance between the two vectors. We expect this \emph{Centre Cosine distance (CCD)} to be minimized. 

For smoothness, the appearance change should be gradual, without abrupt changes such as seams and cuts. 
To measure such, we train a {\em seam classifier} using real samples from the training set as negative examples, and where we create synthetic seams by concatenating two random textures as positive examples. We run this classifier on the center crop. We call this the \emph{Center Seam Score (CSS)}. The architecture and training details of seam classifier are the same as those of $D^{\mathrm{rec}}$ and $D^{\mathrm{itp}}$. 

{\bf Interpolation realism}. The texture should also look realistic, like the training set. To measure this, we chose the Inception Score~\cite{salimans2016improved} and Sliced Wasserstein Distance (SWD)~\cite{karras2017progressive}, and apply them on the center crops. This gives \emph{Center Inception Score (CIS)} and \emph{Center SWD}, respectively. For \emph{CIS}, we use the state-of-the-art \emph{Inception-ResNet-v2} inception model architecture \cite{szegedy2017inception} finetuned with our two datasets separately.

We also found these metrics do not capture undesired repetitions, a common texture synthesis artifact. We, therefore, trained a {\em repetition classifier} for this purpose. We call this the \emph{Center Repetition Score (CRS)}. The architecture and training details of repetition classifier are almost the same as those of the seam classifier except the input image size is $128 \times 256$ instead of $128 \times 128$, where the negative examples are random crops of size $128 \times 256$ from real datasets and the positive examples are horizontally tiled twice from random crops of size $128 \times 128$ from real datasets.

\begin{figure}
\centering
\includegraphics[width=1\linewidth]{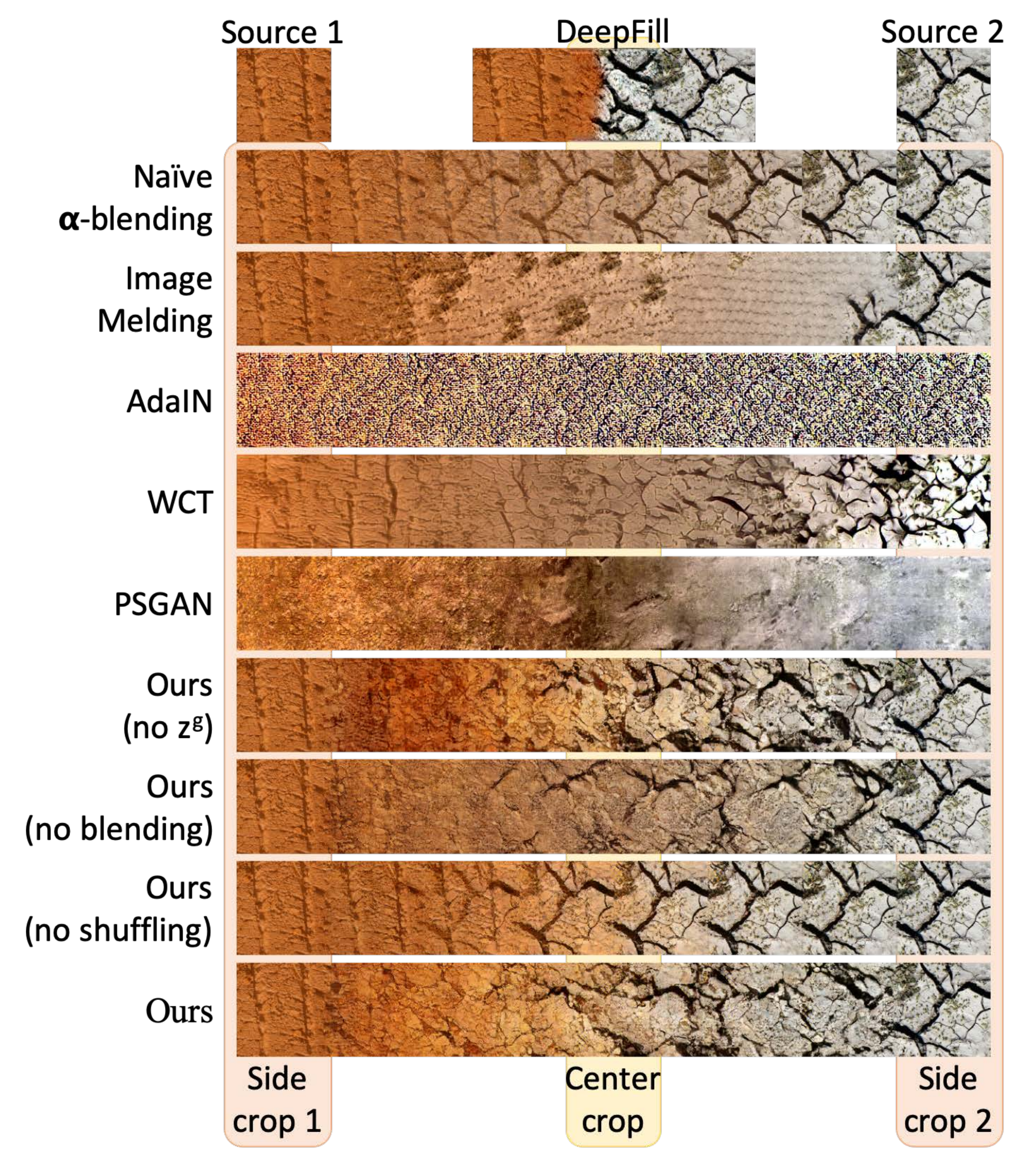}
\caption{Qualitative demonstrations and comparisons of horizontal interpolation in the size of $128 \times 1024$ on the \emph{earth texture} samples. We use the two side crops with the orange background for SPD measurement, and the center crop with the light yellow background for the other proposed quantitative evaluations. For the DeepFill \cite{yu2018generative} method, since the default design is not suitable for inpainting a wide hole due to lack of such ground truth, we instead test it on a shorter interpolation of size $128 \times 384$.}
\label{fig:qual_eval}
\vspace{-12pt}
\end{figure}

\begin{table*}[!t]
\center
\small
\caption{Quantitative evaluation averaging over the \emph{earth texture} and \emph{animal texture} datasets. We highlighited the \textbf{best}, \underline{second best} and \textcolor{red}{very high} values for each metric. We also indicate for each whether higher ($\Uparrow$) or lower ($\Downarrow$) values are more desirable.
}
\begin{tabular}{L{2.5cm}||C{0.9cm}|C{0.8cm}||C{0.8cm}|C{0.9cm}||C{0.9cm}|C{0.8cm}|C{1.0cm}||C{0.8cm}|C{1.1cm}||C{1.0cm}}
\hline
& \multicolumn{2}{c||}{Controllability} & \multicolumn{2}{c||}{Smoothness} & \multicolumn{3}{c||}{Realism} & \multicolumn{2}{c||}{User study} & Testing \tabularnewline
\cline{2-10}
& SPD & CGD & CCD & CSS & CRS & CIS & CSWD & PR & p-value & time\tabularnewline
& $\Downarrow$ & $\Downarrow$ & $\Downarrow$ & $\Downarrow$ & $\Downarrow$ & $\Uparrow$ & $\Downarrow$ & & \tabularnewline
\hline\hline
Na\"{\i}ve $\alpha$-blending & 0.0000 & 1.255 & 0.777 & \textcolor{red}{0.9953} & \textcolor{red}{0.4384} & 22.35 
 & 60.93 & 0.845 & $< 10^{-6}$ & 0.02 s \tabularnewline
Image Melding \cite{darabi2012image} & 0.0111 & 1.289 & 0.865 & 0.0005 & 0.0004 & 
\textbf{29.45} & 47.09 & 0.672 & $< 10^{-6}$ & 6 min\tabularnewline
WCT \cite{li2017universal} & \textcolor{red}{0.8605} & 1.321 & 0.988 & 0.0020 & 0.0000 & 9.86 & 46.89 & 0.845 & $< 10^{-6}$ & 7.5 s\tabularnewline 
PSGAN \cite{pmlr-v70-bergmann17a} & \textcolor{red}{1.1537} & 1.535 & 1.156 & 0.0069 & 0.0005 & \underline{26.81} & 35.90 & 0.967 & $< 10^{-6}$ & 1.4 min \tabularnewline
\hline
Ours (no $z^g$) & 0.0112 & 1.207 & 0.680 & 0.0078 & 0.0010 & 21.04 & \underline{21.54} & - & - & -\tabularnewline
Ours (no blending) & 0.0103 & 1.272 & 0.817 & 0.0125 & 0.0009 & 22.24 & 52.29 & - & - & -\tabularnewline
Ours (no shuffling) & 0.0107 & \textbf{1.129} & \textbf{0.490} & \textcolor{red}{0.0534} & \textcolor{red}{0.2386} & 26.78 & \textbf{20.99} & - & - & -\tabularnewline
Ours & 0.0113 & \underline{1.177} & \underline{0.623} & 0.0066 & 0.0008 & 26.68 & 22.10 & - & - & 0.5 s\tabularnewline
\hline
\end{tabular}
\label{quant_eval}
\vspace{-12pt}
\end{table*}

\begin{figure*}[!t]
\centering
\includegraphics[width=1\linewidth]{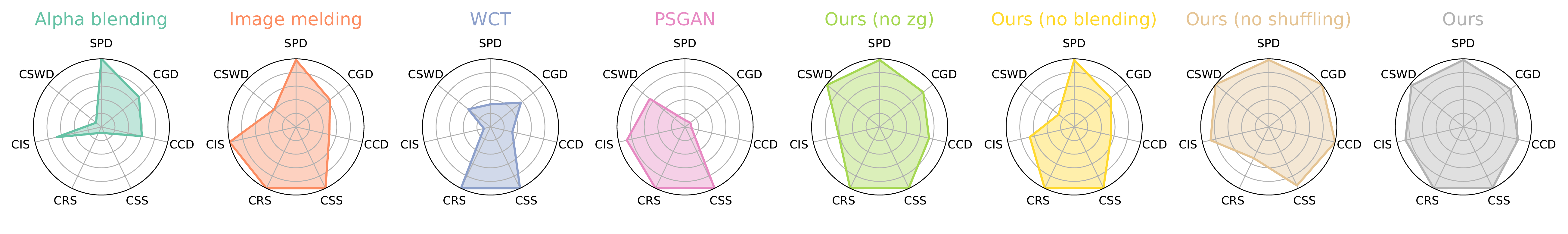}
\caption{Radar charts visualizing Table~\ref{quant_eval}. Values have been normalized to the unit range, and axes inverted so that higher value is always better. The first four are baseline methods and next three ablation candidates, with the last entry representing our full method. Our method scores near-top marks all around and shows balanced performance according to all metrics.
}
\label{fig:radar_plot}
\vspace{-12pt}
\end{figure*}

\subsection{Comparisons}
\label{sec:comparisons}
We compare against several leading methods from different categories on the task of texture interpolation. These include:  na\"{\i}ve $\alpha$-blending, Image Melding~\cite{darabi2012image} as a representative of patch-based techniques, two neural stylization methods - AdaIN~\cite{huang2017arbitrary} and WCT~\cite{li2017universal}, a recent deep hole-filling method called DeepFill~\cite{yu2018generative}, and PSGAN \cite{pmlr-v70-bergmann17a} which is the closest to ours but without user control. Most these had to be adapted for our task. See more details in the supplementary material. Fig.~\ref{fig:qual_eval} contains a qualitative comparison between the different methods. Note that in this example: (1) the overly sharp interpolation of DeepFill, (2) the undesired ghosting and repetition artifacts of na\"{\i}ve $\alpha$-blending and ours (no shuffling), (3) the incorrect reconstruction and less relevant interpolation of AdaIN, WCT, and PSGAN, (4) the appearance mismatch between source and interpolation of Image Melding, (5) the lack of smoothness of ours (no $z^g$), and (6) the undesired fading of ours (no blending). More qualitative comparisons are shown in the supplementary material. We also report qualitative results, including the user study and the ablation experiments, in Table~\ref{quant_eval}, that contains average values for the two datasets - \emph{earth texture} and \emph{animal texture}. \fig{fig:radar_plot} summarizes the quantitative comparisons.

\subsection{User study}
\label{sec:user_study}
We also conducted a user study on Amazon Mechanical Turk. We presented the users with a binary choice, asking them if they aesthetically prefer our method or one of the baseline methods on a random example from the horizontal interpolation task. The user study webpage and sanity check (to guarantee the effectiveness of users' feedback) are shown in the supplementary material. For each method pair, we sampled $90$ examples and collected $5$ independent user responses per example. Tallying the user votes, we get $90$ results per method pair. We assumed a null hypothesis that on average, our method will be preferred by $2.5$ users for a given method pair. We used a one-sample permutation t-test to measure p-values, using $10^6$ permutations, and found the p-values for the null hypothesis are all $< 10^{-6}$. This indicates that the users do prefer one method over another. To quantify this preference, we count for each method pair all the examples where at least $3$ users agree in their preference, and report a \emph{preference rate (PR)} which shows how many of the preferences were in our method's favor. Both PR and the p-values are listed in Table~\ref{quant_eval}.

\subsection{Ablation study}
\label{sec:ablation_study}
We also compare against simplified versions of our method. The qualitative results for this comparison are shown in \fig{fig:qual_eval}. We report quantitative result numbers in Table~\ref{quant_eval}, and visualized them in \fig{fig:radar_plot}. We ablate the following components:

{\bf Remove $\mathbf{z^g}$}. The only difference between $z^g$ and $z^l$ is in the tiling and shuffling for $z^l$. However, if we remove $z^g$, we find texture transitions are less smooth and gradual.

{\bf Remove texture blending during training}. We modify our method so that the interpolation task during training is performed only upon two identical textures. This makes the interpolation discriminator $D^{\mathrm{itp}}$ not be aware of the realism of blended samples, so testing realism deteriorates.

{\bf Remove random shuffling}. We skip the shuffling operation in latent space and only perform blending during training. This slightly improves realism and interpolation directness, but causes visually disturbing repetitions.

\vshrink{}
\section{Conclusion}
We presented a novel method for controllable interpolation of textures. We were able to satisfy the criteria of controllability, smoothness, and realism. Our method outperforms several baselines on our newly collected datasets. As we see in \fig{fig:radar_plot}, although some baseline method may achieve better results than ours on one of the evaluation criteria, they usually fail on the others. In contrast, our method has consistent high marks in all evaluation categories. The user study also shows the users overwhelmingly prefer our method to any of the baselines. We have also demonstrated several applications based on this technique and hope it may become a building block of more complex workflows.

\vshrink{}
\section*{Acknowledgement}
\ning{The authors acknowledge the Maryland Advanced Research Computing Center for providing computing resources and acknowledge the photographers for licensing photos under Creative Commons or public domain.}

{\small
\bibliographystyle{ieee}
\bibliography{CVPRreferencelist}
}

\clearpage
\renewcommand\thesubsection{\Alph{subsection}}

\section{Supplementary material}

\subsection{Shuffling procedure visualization}
\begin{figure}
\centering
\includegraphics[width=1\linewidth]{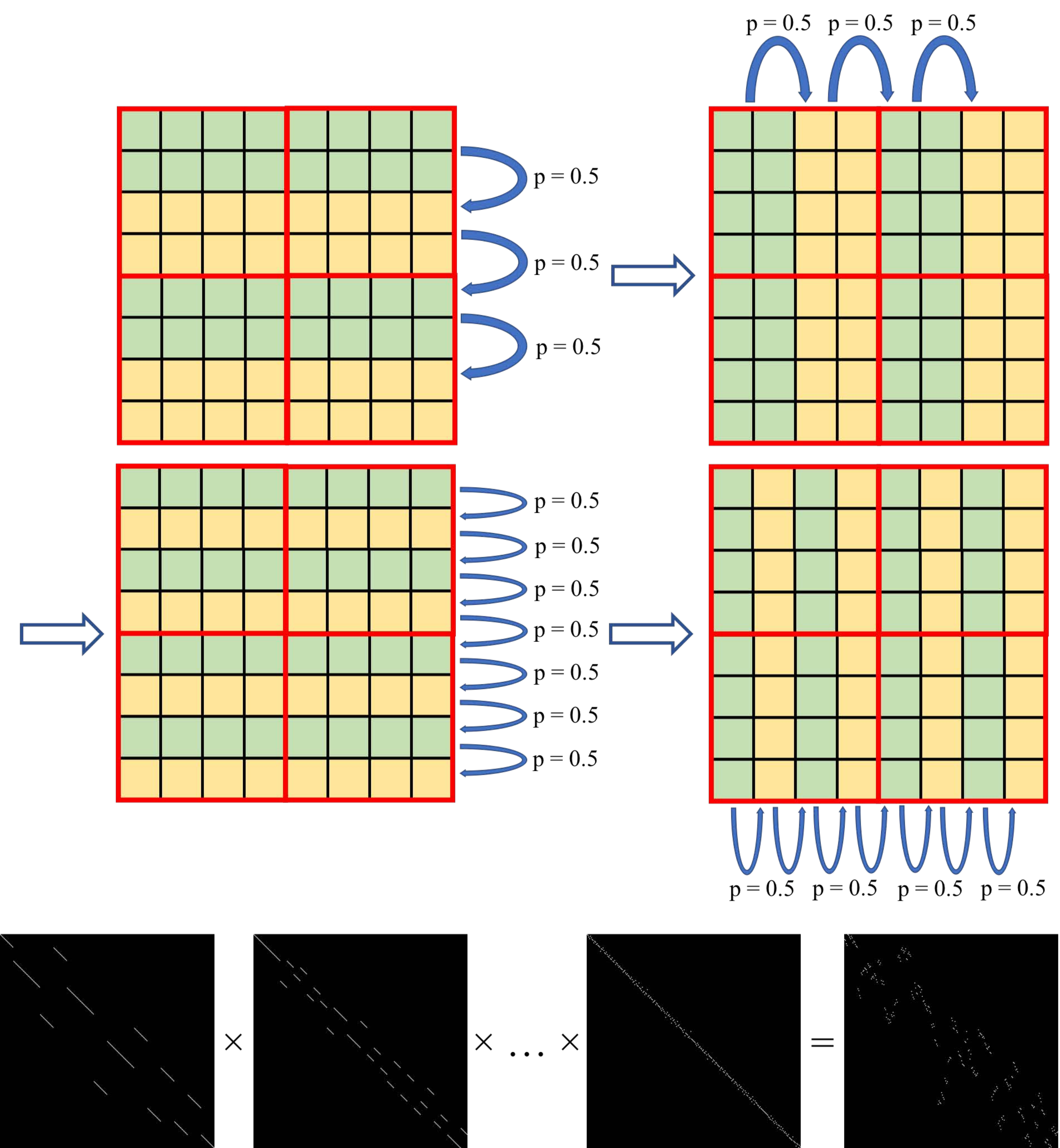}
\caption{Our shuffling procedure. On the top figure we visualize the procedure example $P_{0}\circ P_{1}(T(z^l_i))$, where $z^l_i$ is a $4 \times 4$ cell in a $2 \times 2$ grid framed in red. The procedure is composed of the random swapping operations between a green strip and its subsequently adjacent yellow strip in four directions: top-down, bottom-up, left-to-right, and right-to-left. The swapping operations start at scale 2 (the 1st row) and then are repeated at scale 1 (the 2nd row). The bottom figure (zoom-in to check) demonstrates the composition of swapping operations at several scales applied to an identity matrix. The resulting composed matrix can serve as a row permutation matrix left-multiplied to $T(z_i^l)$. Another similar matrix can serve as a column permutation matrix right-multiplied to $T(z_i^l)$. The row and column permutation matrices are independently sampled for each training iteration.}
\label{fig:shuffling}
\end{figure}

We visualize our shuffling procedure in \fig{fig:shuffling}.

\begin{figure}
\centering
\includegraphics[width=3.25in]{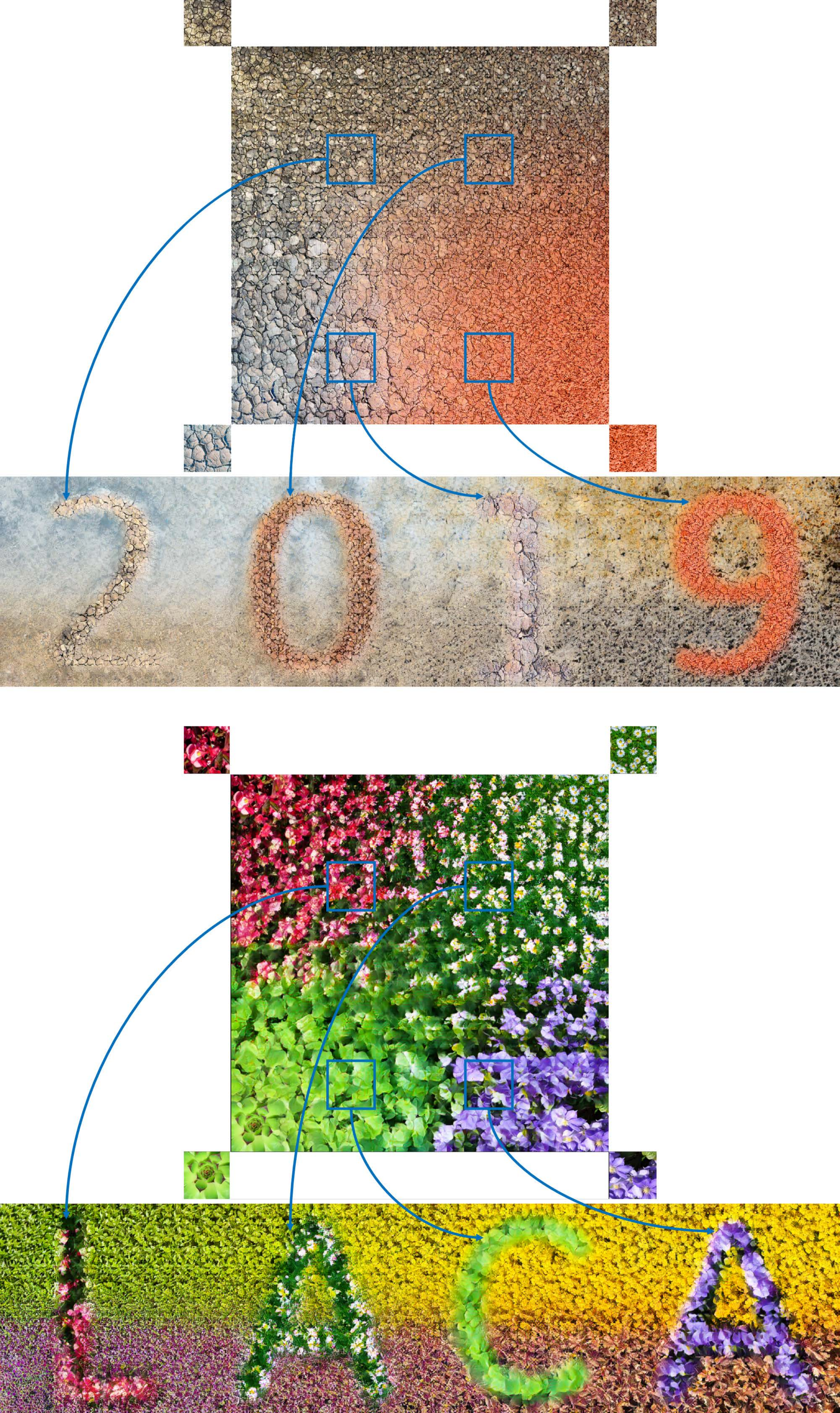}
\caption{Texture interpolation and texture painting using our network on the \emph{earth texture} and \emph{plant texture} datasets. The top part shows a $1024 \times 1024$ palette created by interpolating four source textures at the corners outside the palette. The bottom part shows a $512 \times 2048$ painting of letters with different textures sampled from the palette. The letters are interpolated by our method with the background, also generated by our interpolation.} 
\label{fig:palette_brush_earth_plant}
\end{figure}

\begin{figure}
\centering
\includegraphics[width=3.25in]{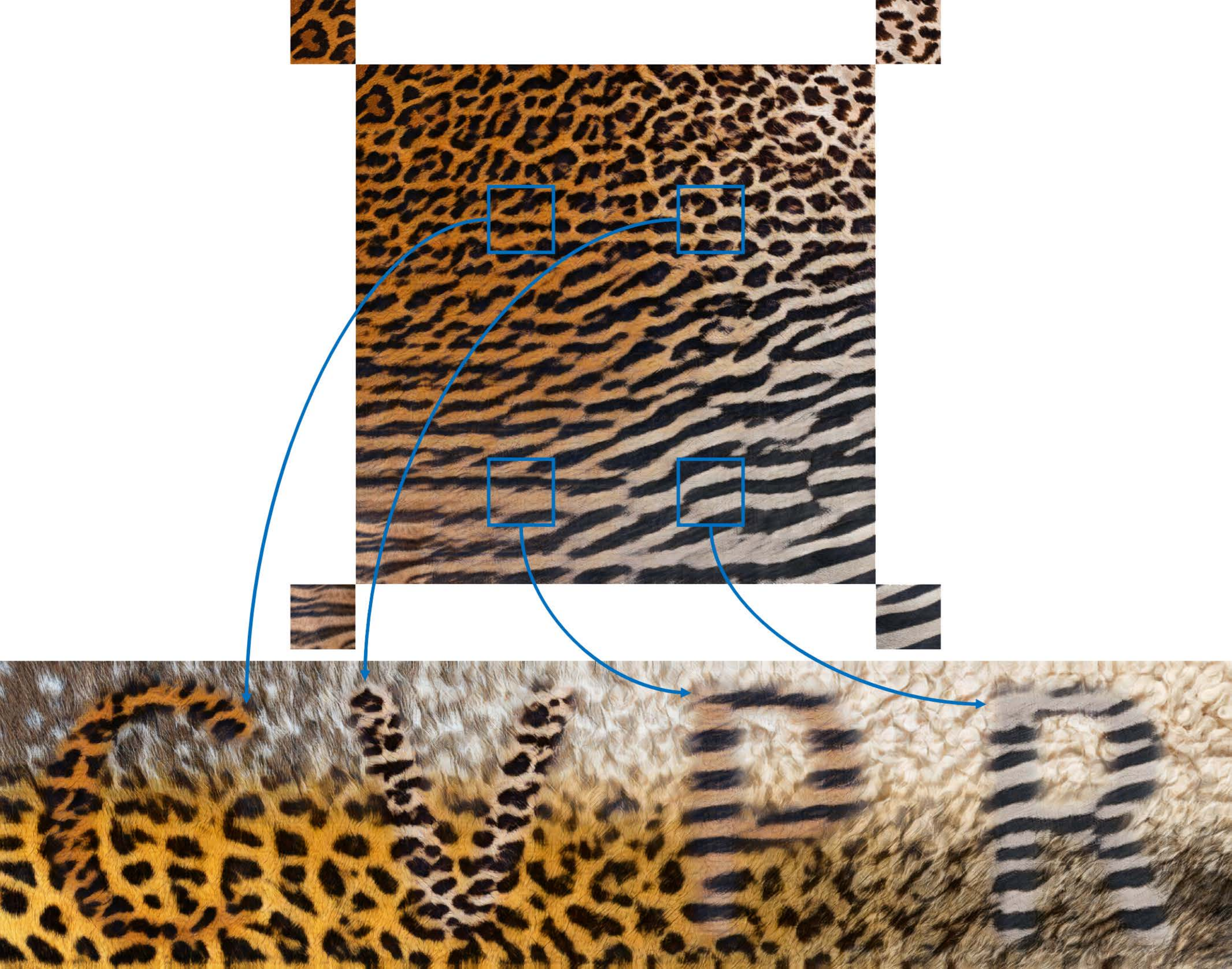}
\caption{Texture interpolation and texture painting with camouflage effect using our network on the \emph{animal texture} dataset. The top part shows a $1024 \times 1024$ palette created by interpolating four source textures at the corners outside the palette. The bottom part shows a $512 \times 2048$ painting of letters with different textures sampled from the palette. The letters are interpolated by our method with the background, also generated by our interpolation.} 
\label{fig:palette_brush_animal_camouflage}
\end{figure}

\subsection{Texture palette and brush examples}
In order to diversify our applications, we, in addition, collected a \emph{plant texture} dataset from Adobe Stock and randomly split it into $1,074$ training and $119$ testing images. We show the texture palette and brush application on the \emph{earth texture} and \emph{plant texture} datasets in \fig{fig:palette_brush_earth_plant}. Furthermore, we show in \fig{fig:palette_brush_animal_camouflage} a camouflage effect of brush painting on the \emph{animal texture} dataset, intentionally given the background patterns similar to brush patterns. It indicates the smooth interpolation over different textures. The dynamic processes of drawing such paintings plus the painting of Figure~1 in the main paper are demonstrated in the videos at \href{https://github.com/ningyu1991/TextureMixer.git}{GitHub}. The videos are encoded using MP4 libx265 codec at $60$ frame rate and $16$M bit rate.

\begin{figure*}
\centering
\includegraphics[width=1\linewidth]{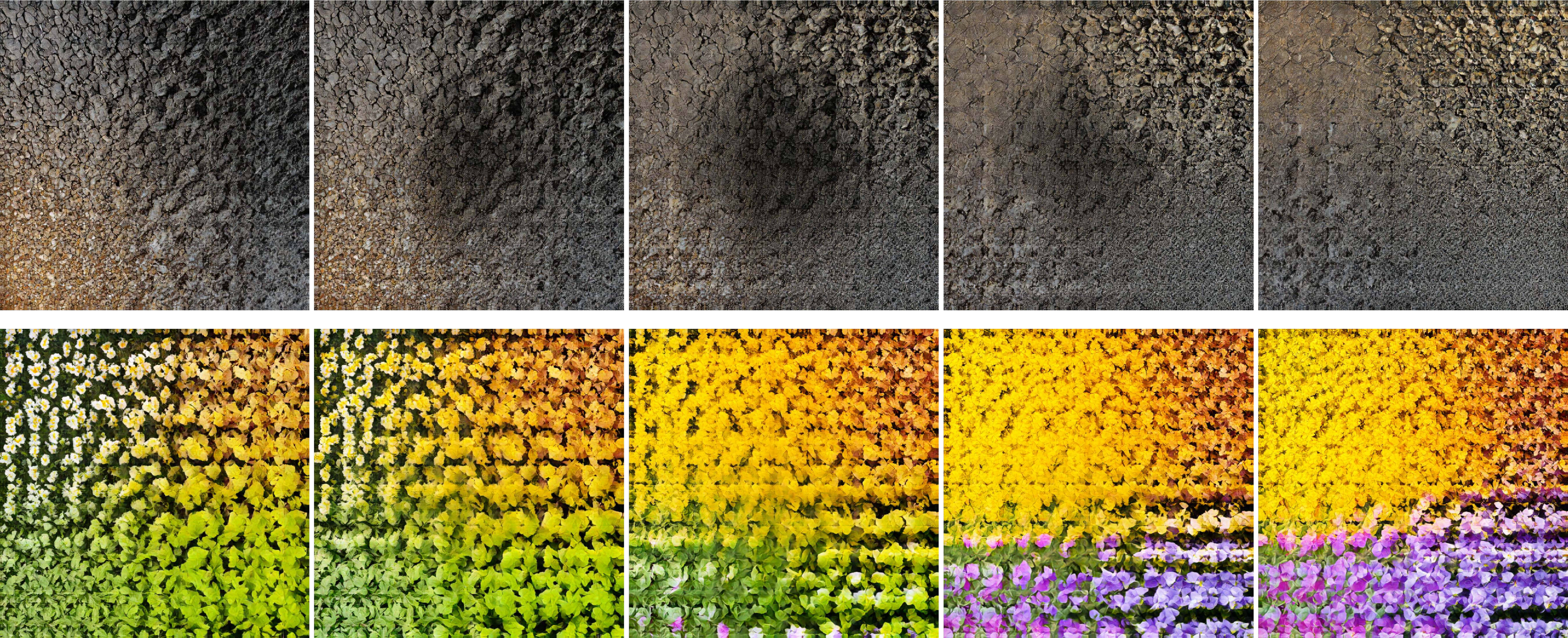}
\caption{Sequences of dissolve video frame samples with size $1024 \times 1024$ on the \emph{earth texture} and \emph{plant texture} datasets, where each frame is also with effect of interpolation.}
\label{fig:dissolve_earth_plant}
\end{figure*}

\subsection{Texture dissolve examples}
We shows in \fig{fig:dissolve_earth_plant} additional sequences of video frame samples with gradually varying weights on the \emph{earth texture} and \emph{plant texture} datasets. The corresponding videos plus the video for Figure~3 in the main paper are at \href{https://github.com/ningyu1991/TextureMixer.git}{GitHub}. The videos are encoded using MP4 libx265 codec at $60$ frame rate and $16$M bit rate.

\subsection{Animal hybridization details and examples}
\begin{figure*}
\centering
\includegraphics[width=1\linewidth]{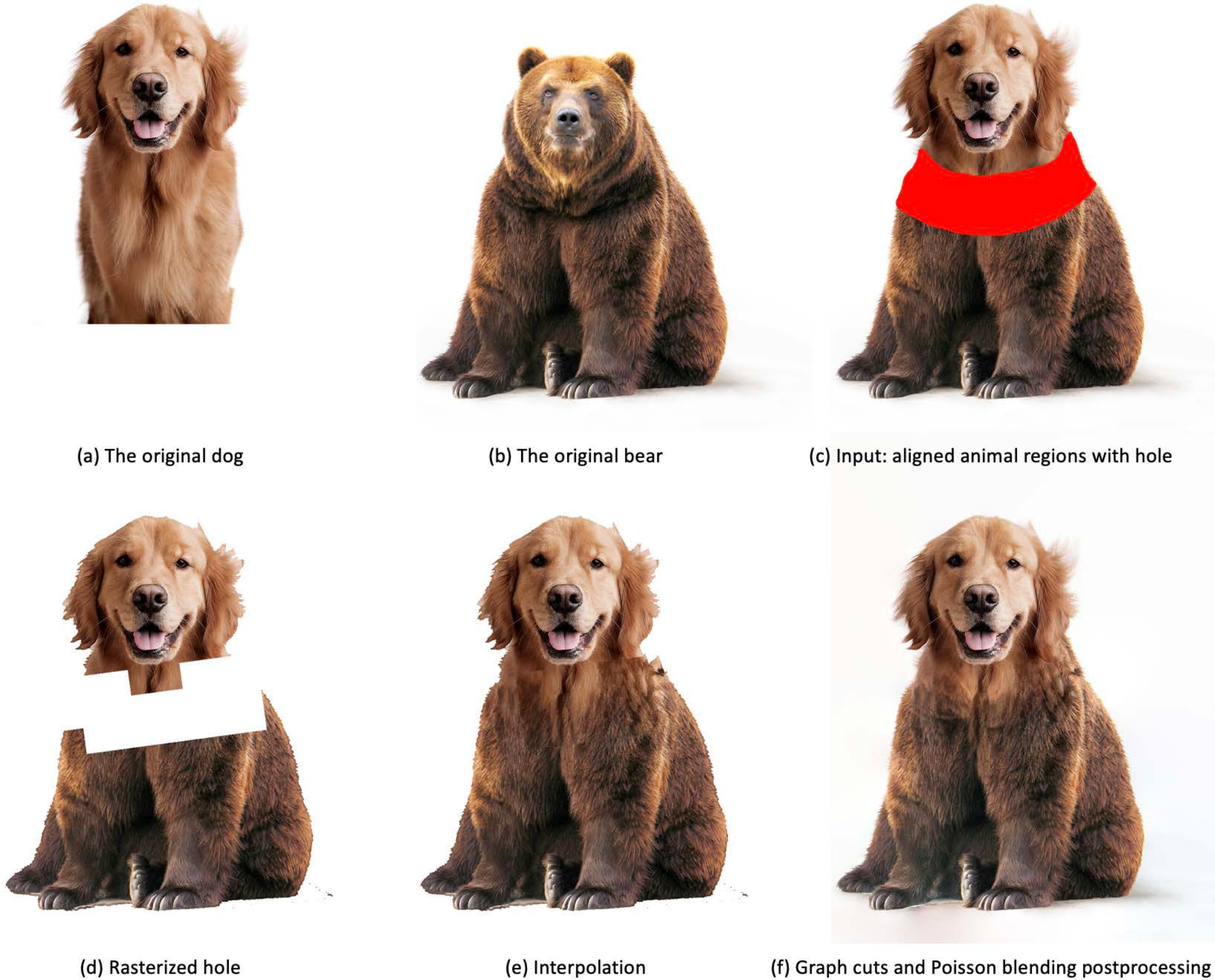}
\caption{Animal hybridization pipeline. (a) and (b) are two original images. (c) is the input to the pipeline, composed of the aligned regions of (a) and (b) in the same image and the hole for the transition region. (d) shows that we rasterize the hole because Texture Mixer works on square patches. The patch size is $128 \times 128$. (e) shows that we interpolate in the rasterized hole region using adjacent texture patches, and then composite this back on top of the original image. This involves two details: (1) if a texture patch covers background, those background pixels are replaced by foreground pixels using the Content-Aware Fill function in Photoshop; and (2) we blend latent tensors between two images using spatially varying weights. (f) We use graph cuts \cite{kwatra2003graphcut} and standard Poisson blending \cite{perez2003poisson} to postprocess the boundaries.}
\label{fig:beardog_pipeline} 
\end{figure*}

\begin{figure*}
\centering
\includegraphics[width=1\linewidth]{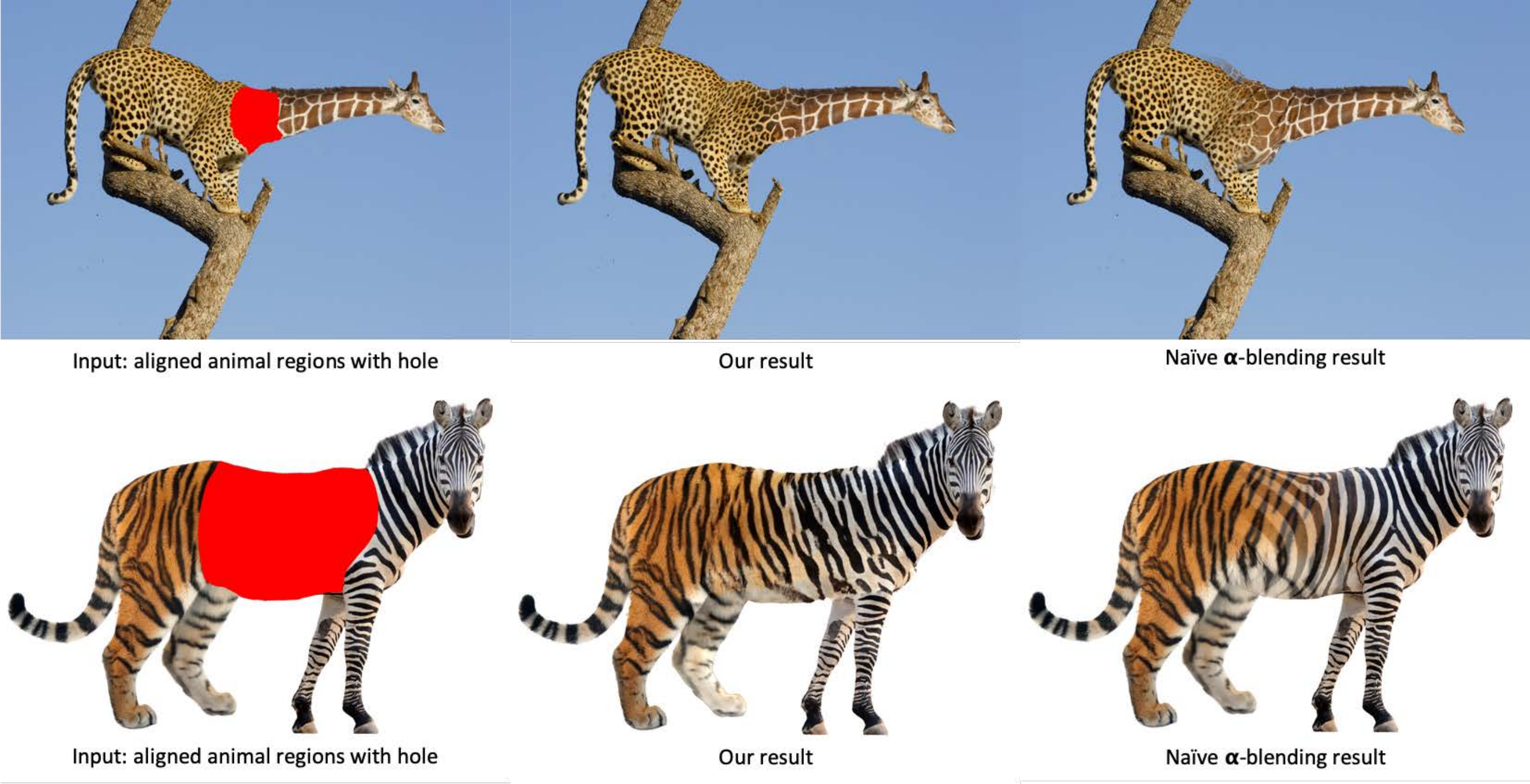}
\caption{Two animal hybridization examples. The top image is in the size $2636 \times 3954$ and the bottom image is in the size $2315 \times 2664$. Our interpolations between the two animal furs is smoother, has less ghosting, and is more realistic than those of the Na\"{\i}ve $\alpha$-blending.}
\label{fig:leoraffe_ziger_hybridization}
\end{figure*}

In \fig{fig:beardog_pipeline}, we show and illustrate the pipeline to hybridize a dog and a bear by interpolating their furs in the hole for the transition region. Two additional results are shown in \fig{fig:leoraffe_ziger_hybridization}.

\subsection{Network architecture details}
\label{sec:arch}
We set the texture image size to be $128$ throughout our experiments. The proposed $E^l$, $E^g$, $D^{rec}$, and $D^{itp}$ architectures are employed or adapted from the discriminator architecture in \cite{karras2017progressive}, where layers with spatial resolutions higher than $128 \times 128$ are removed. We also adopt their techniques including \emph{pixel normalization} instead of batch normalization, and \emph{leaky ReLU activation}. The \emph{minibatch standard deviation channel} is also preserved for $D^{rec}$ and $D^{itp}$, but not for $E^l$ and $E^g$. For $E^l$, we truncate the architecture so that the output local latent tensor is $m$ times smaller than the input texture, where $m=4$ in all our experiments. We tried using deeper architectures but noticed this does not favor reconstruction quality. For $E^g$, we truncate the architecture at $1 \times 1$ resolution right before the fully-connected layer, because we are doing encoding rather than binary classification.

Our $G$ is modified from the fully-convolutional generator architecture from Karras~\etal~\cite{karras2017progressive} with three changes. First, the architecture is truncated to accept an input spatial resolution that is $m=4$ times smaller than the texture size, and to output the original texture size. Second, the local and global latent tensor inputs are concatenated together along the channel dimension after they are fed into $G$. A third important point is that since our goal is to interpolate a larger texture image output, at the bottleneck layer the receptive field should be large enough to cover the size of input image. We do this by inserting a chain of five residual blocks \cite{he2016deep} in the generator after local and global latent tensor concatenation and before the deconvolution layers from \cite{karras2017progressive}.

\subsection{Training details}
\label{sec:implementation}
Our training procedure again follows the progressive growing training in \cite{karras2017progressive}, where $E^l$, $E^g$, $G$, $D^{rec}$, and $D^{itp}$ simultaneously grow from image spatial resolution at $32 \times 32$ to $128 \times 128$. We repeatedly alternate between performing one training iteration on $D^{rec}$ and $D^{itp}$, and then four training iterations on $E^l$, $E^g$, and $G$. At each intermediate resolution during growth, the \emph{stabilization stage} takes 1 epoch of training and the \emph{transition stage} takes 3 epochs. After the growth is completed, we keep training the model until a total of 20 epochs is reached.

We use Adam \cite{kingma2014adam} as our optimization approach with no exponential decay rate $\beta_1 = 0.0$ for the first moment estimates and with the exponential decay rate for the second moment estimates $\beta_2 = 0.99$. The learning rate is set to $0.001$ before the model grows to the final resolution $128 \times 128$ and then is set to $0.0015$ at $128 \times 128$. The trainable weights of the autoencoder and discriminator are initialized with the \emph{equalized learning rate} technique from \cite{karras2017progressive}. We train and test all our models on 8 NVIDIA GeForce GTX 1080 Ti GPUs with 12GB of GPU memory each. Based on the memory available and the training performance, we set the batch size at $64$, and the training lasts for 3 days.

The weights of losses is not sensitive to the dataset. We simply set them to balance the order of magnitude of each loss: $\lambda_1 = 100$, $\lambda_2 = \lambda_4 = 0.001$, and $\lambda_3 = \lambda_5 = 1$.


\subsection{Experimental evaluation details}
{\bf Seam classifier}. The architecture and training details of seam classifier are almost the same as those of $D^{rec}$ and $D^{itp}$ except (1) we remove the \textit{minibatch standard deviation channel}, (2) we add a sigmoid activation layer after the output layer for the binary cross-entropy loss computation, and (3) we exclude the progressive growing process. We directly use the sigmoid output of the classifier as the seam score for each input image.

{\bf Repetition classifier}. The architecture and training details of repetition classifier are almost the same as those of the seam classifier except the input image size is $128 \times 256$ instead of $128 \times 128$, where the negative examples are random crops of size $128 \times 256$ from real datasets and the positive examples are horizontally tiled twice from random crops of size $128 \times 128$ from real datasets.

{\bf Inception model finetuning}. Our inception scores are computed from the state-of-the-art \emph{Inception-ResNet-v2} inception model architecture \cite{szegedy2017inception} finetuned with our two datasets separately.

\subsection{Baseline method details}
{\bf Na\"{\i}ve $\mathbf{\alpha}$-blending}. We split the output into 8 square tiles, where the end textures are copied as-is, and the intervening tiles (copies of the two boundaries) are linearly per-pixel $\alpha$-blended.

{\bf Image Melding \cite{darabi2012image}}. We selected Image Melding in its inpainting mode as a representative of patch-based methods. We use the default setting of the official public implementation\footnote{\url{https://www.ece.ucsb.edu/~psen/melding}}.

{\bf AdaIN \cite{huang2017arbitrary}}. Style transfer techniques can potentially be leveraged for the interpolation task by using random noise as the content image and texture sample as the style. We interpolate the neural features of the two source textures to vary the style from left to right. We consider AdaIN as one representative of this family of techniques, as it can run with arbitrary content and style images. However, with the default setting of the official implementation\footnote{\url{https://github.com/xunhuang1995/AdaIN-style}} and their pre-trained model, AdaIN has some systematic artifacts as it over-preserves the noise appearance. Therefore, we only show qualitative results in Figure~4 in the main paper, and in \fig{fig:qual_eval_2} to \fig{fig:qual_eval_6} here. We did not include this method in the quantitative evaluation.


{\bf WCT \cite{li2017universal}}. WCT is an advancement over AdaIN with whitening and coloring transforms (WCT) as the stylization technique and works better on our data. We use its official public implementation\footnote{\url{https://github.com/Yijunmaverick/UniversalStyleTransfer}} with default setting and their pre-trained model. By design, this method does not guarantee accurate reconstruction of input samples.

{\bf DeepFill \cite{yu2018generative}}. Texture interpolation can be considered an instance of image hole-filling. The training code for the most recent work in this area~\cite{liu2018image} is not released yet. We, therefore, tried another recent method called DeepFill~\cite{yu2018generative} with their official code\footnote{\url{https://github.com/JiahuiYu/generative_inpainting}}. We re-trained it for our two texture datasets separately with $256 \times 256$ input image size, $128 \times 128$ hole size, and all the other default settings. The interpolation results suffered from two major problems: (i) the method is not designed for inpainting wide holes (in our experiment $128 \times 768$) because of lack of such wide ground truth; (ii) even for a smaller hole with size $128 \times 128$, as shown in the failure cases in Figure~4 in the main paper and in \fig{fig:qual_eval_2} to \fig{fig:qual_eval_6}, this work systematically failed to merge the two source textures gradually. We, therefore, excluded this method from our quantitative comparisons.

{\bf PSGAN \cite{pmlr-v70-bergmann17a}}. The most closely related work to ours, PSGAN, learns a smooth and complete neural manifold that favors interpolation. However, it only supports constraining the interpolation in latent space, and lacks a mechanism to specify end texture conditions using image examples. To allow for a comparison, we have trained a PSGAN model for each of our datasets separately, using the official code\footnote{\url{https://github.com/zalandoresearch/psgan}} and default settings. Then, we optimize for the latent code that corresponds to each of the end texture images by backpropagating through L-BFGS-B \cite{byrd1995limited}. We use the gradients of the $L_1$ reconstruction loss and the Gram matrix loss \cite{johnson2016perceptual} and initialize randomly the latent vectors. We use $100$ different initializations and report the best result.


\subsection{More qualitative comparisons}
More qualitative comparisons are shown from \fig{fig:qual_eval_2} to \fig{fig:qual_eval_6}. They are all used for quantitative comparison and user study as reported in Table~1 in the main paper. In addition, \fig{fig:qual_eval_failure} demonstrates one of our failure examples when dealing with strong structural textures.

\begin{figure*}
\centering
\includegraphics[width=1\linewidth]{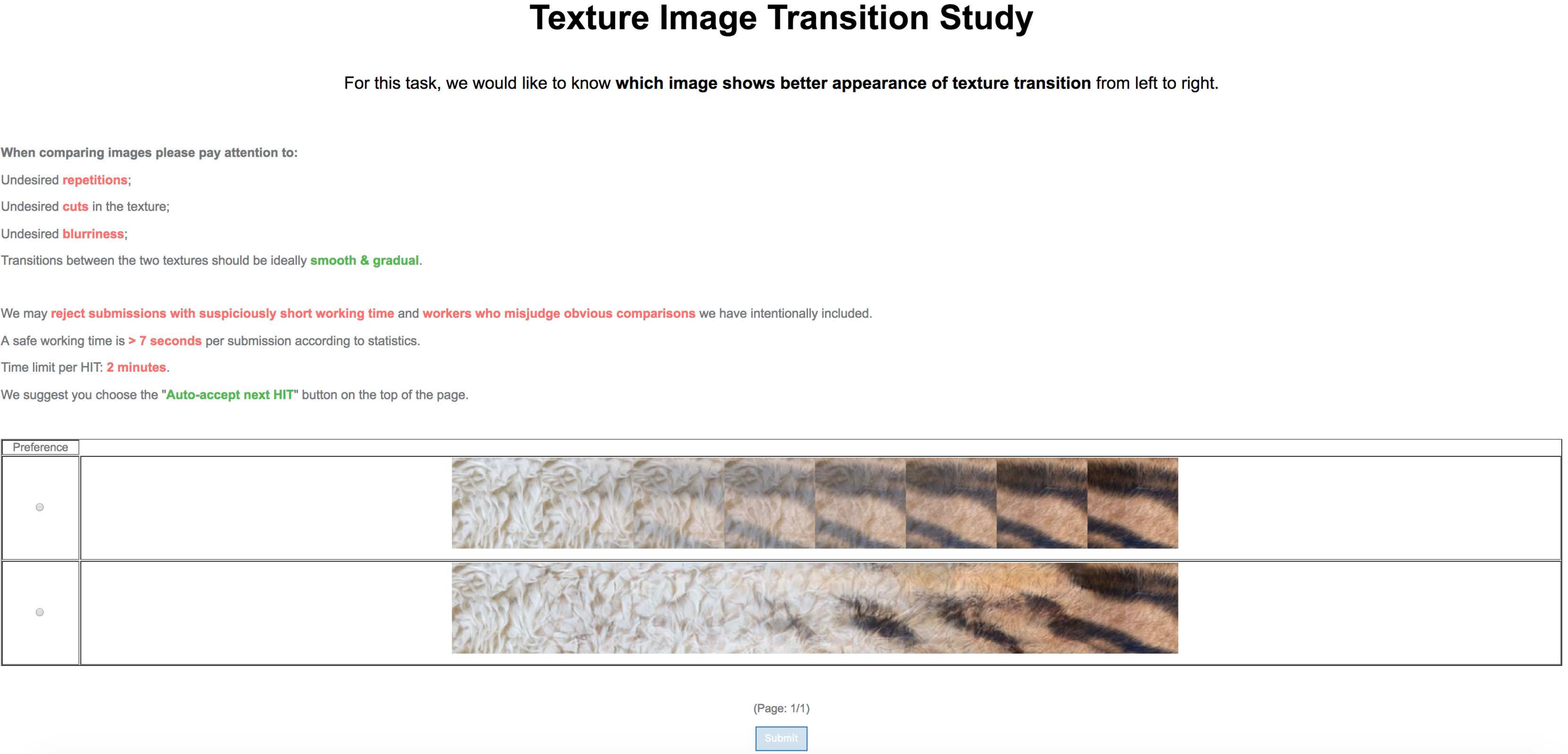}
\caption{User study webpage design.}
\label{fig:user_study_webpage_design}
\end{figure*}

\begin{figure*}
\centering
\includegraphics[width=1\linewidth]{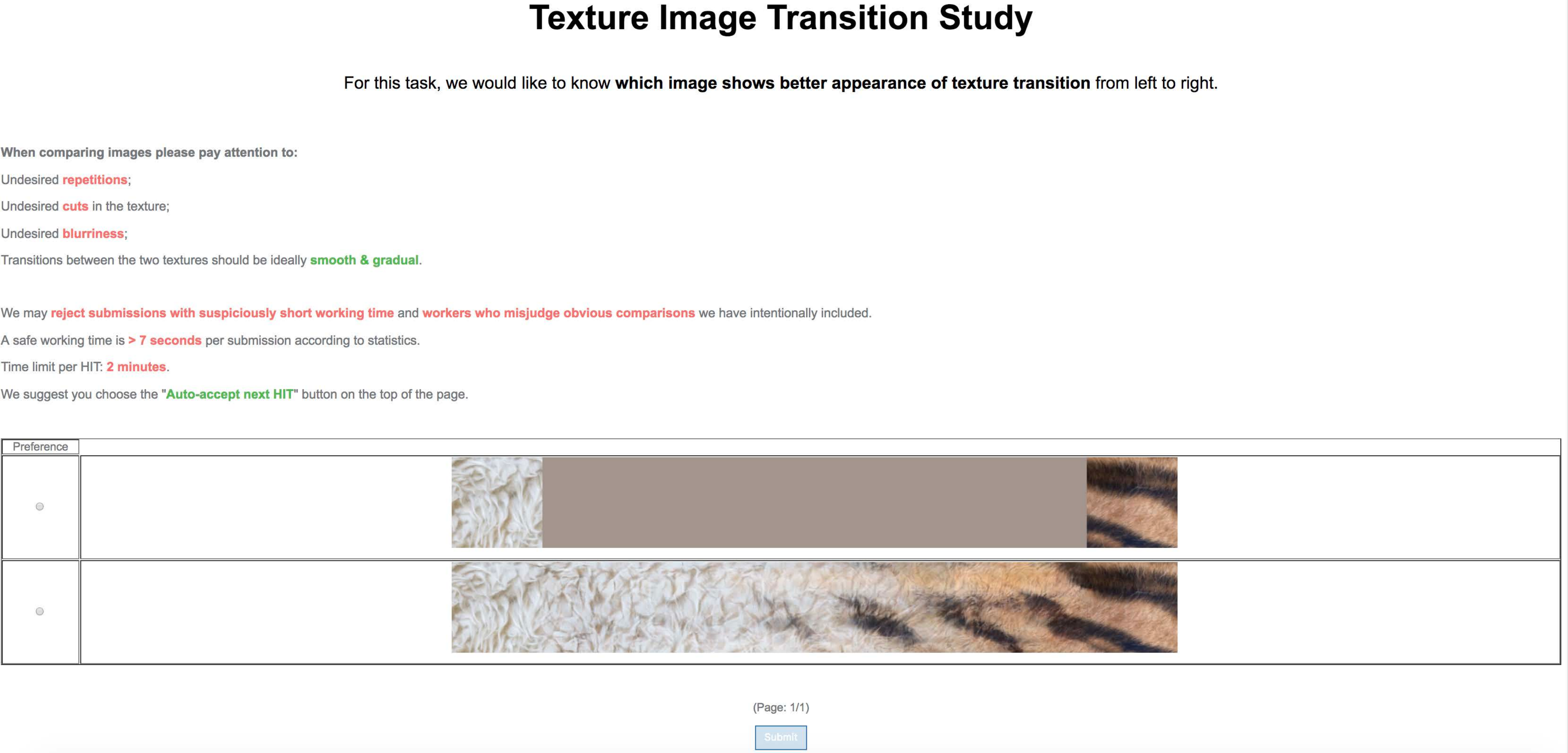}
\caption{A user study with sanity check where the preference should be obvious and deterministic without subjective variance.}
\label{fig:user_study_webpage_design_sanity}
\end{figure*}

\subsection{User study details}
Our user study webpage is shown in \fig{fig:user_study_webpage_design}. To guarantee the accuracy of users' feedback, we insert sanity check by comparing our interpolation results with another naive baseline results where the transition regions are filled with constant pixel values. The constant value is computed as the mean value of the two end texture pixels, as shown in \fig{fig:user_study_webpage_design_sanity}. The preference should be obvious and deterministic without subjective variance. In our statistics, only two users made a mistake once on the sanity check questions. We then manually checked their answers to other real questions but didn't notice any robot or laziness style. We there trust and accept all users' feedback.

\begin{figure*}
\centering
\includegraphics[width=1\linewidth]{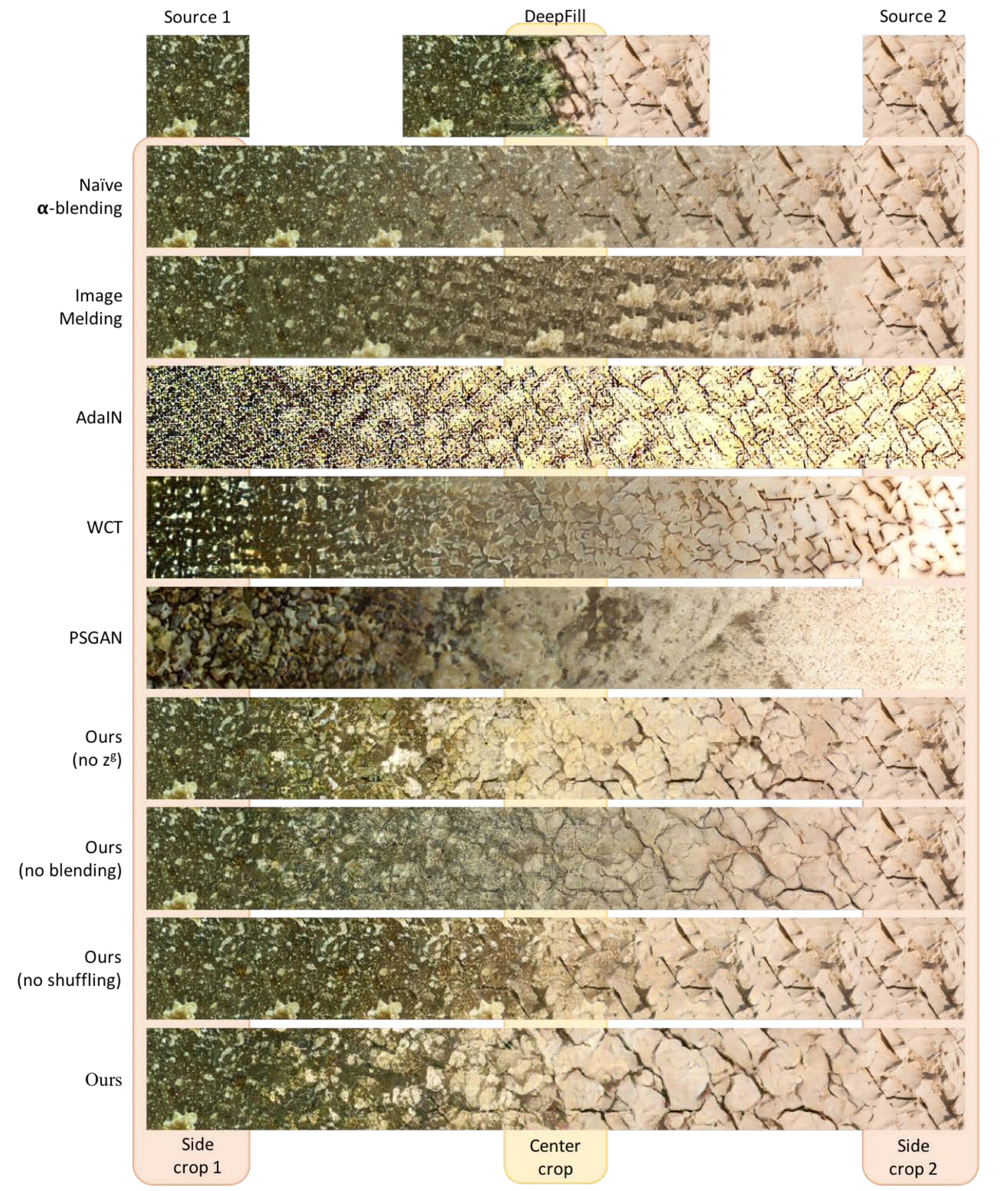}
\caption{Qualitative demonstrations and comparisons of horizontal interpolation in the size of $128 \times 1024$ on the \emph{earth texture} samples. We use the two side crops with the orange background for SPD measurement, and the center crop with the light yellow background for the other proposed quantitative evaluations.}
\label{fig:qual_eval_2}
\end{figure*}

\begin{figure*}
\centering
\includegraphics[width=1\linewidth]{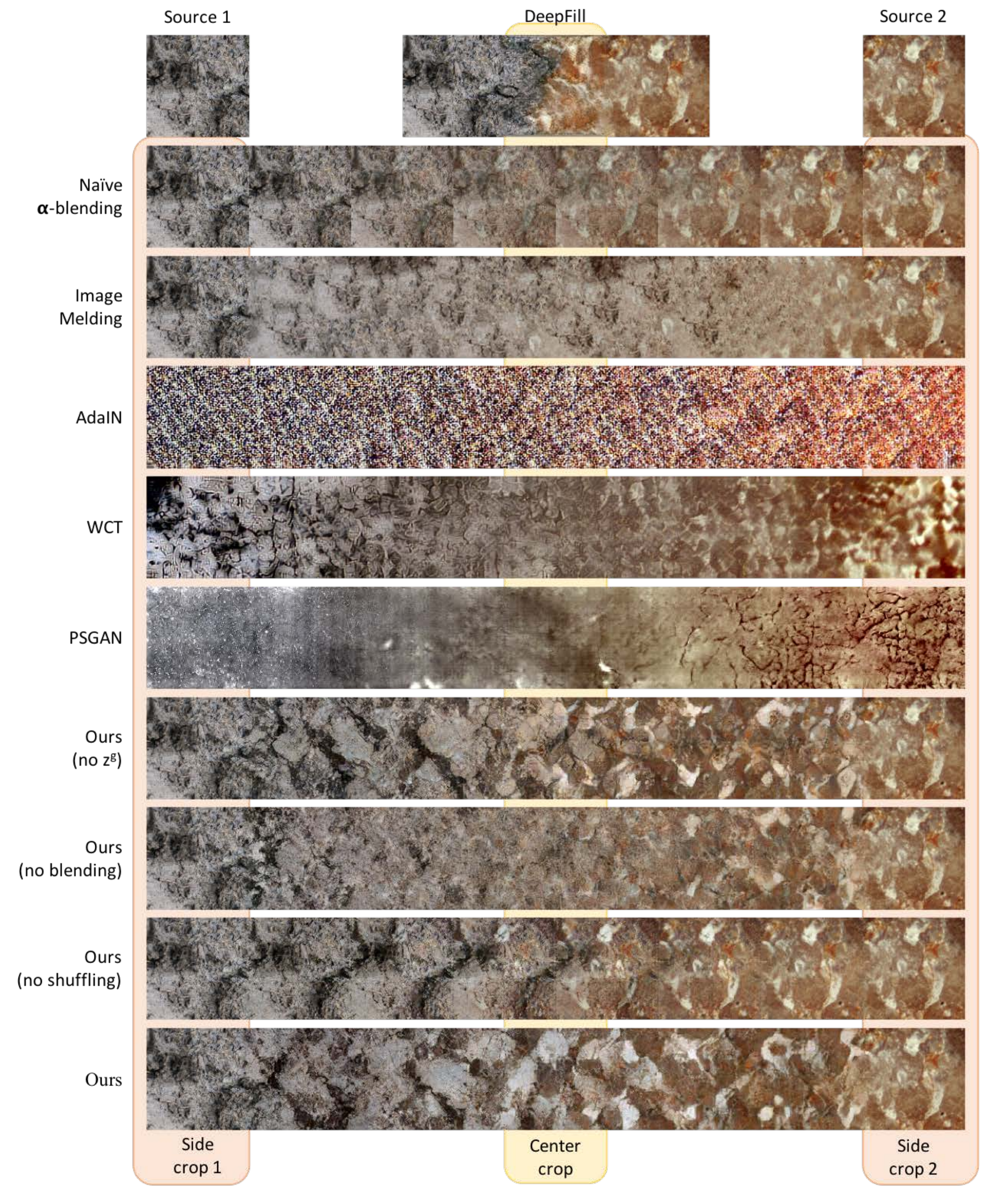}
\caption{Qualitative demonstrations and comparisons of horizontal interpolation in the size of $128 \times 1024$ on the \emph{earth texture} samples. We use the two side crops with the orange background for SPD measurement, and the center crop with the light yellow background for the other proposed quantitative evaluations.}
\label{fig:qual_eval_3}
\end{figure*}

\begin{figure*}
\centering
\includegraphics[width=1\linewidth]{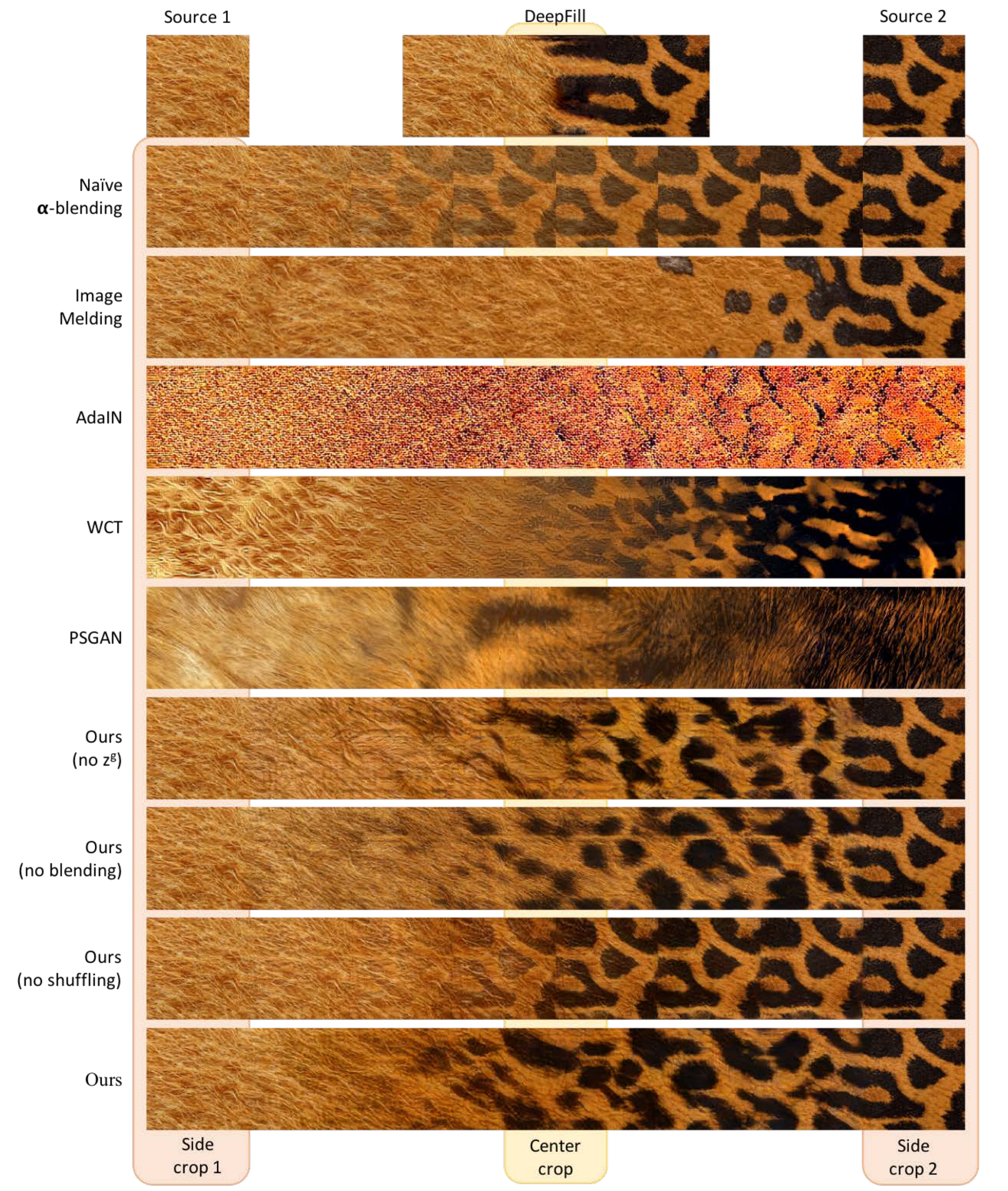}
\caption{Qualitative demonstrations and comparisons of horizontal interpolation in the size of $128 \times 1024$ on the \emph{animal texture} samples. We use the two side crops with the orange background for SPD measurement, and the center crop with the light yellow background for the other proposed quantitative evaluations.}
\label{fig:qual_eval_4}
\end{figure*}

\begin{figure*}
\centering
\includegraphics[width=1\linewidth]{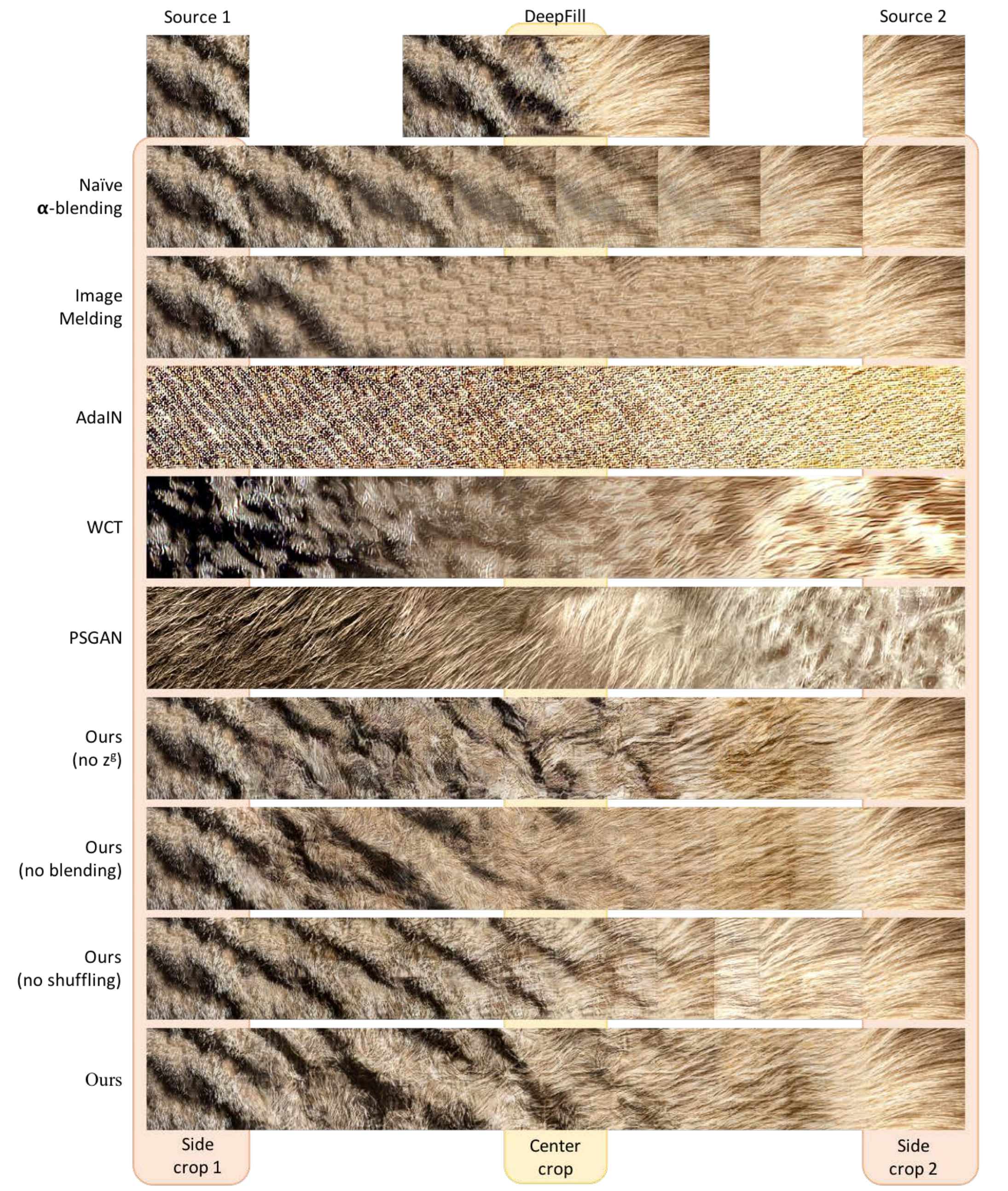}
\caption{Qualitative demonstrations and comparisons of horizontal interpolation in the size of $128 \times 1024$ on the \emph{animal texture} samples. We use the two side crops with the orange background for SPD measurement, and the center crop with the light yellow background for the other proposed quantitative evaluations.}
\label{fig:qual_eval_5}
\end{figure*}

\begin{figure*}
\centering
\includegraphics[width=1\linewidth]{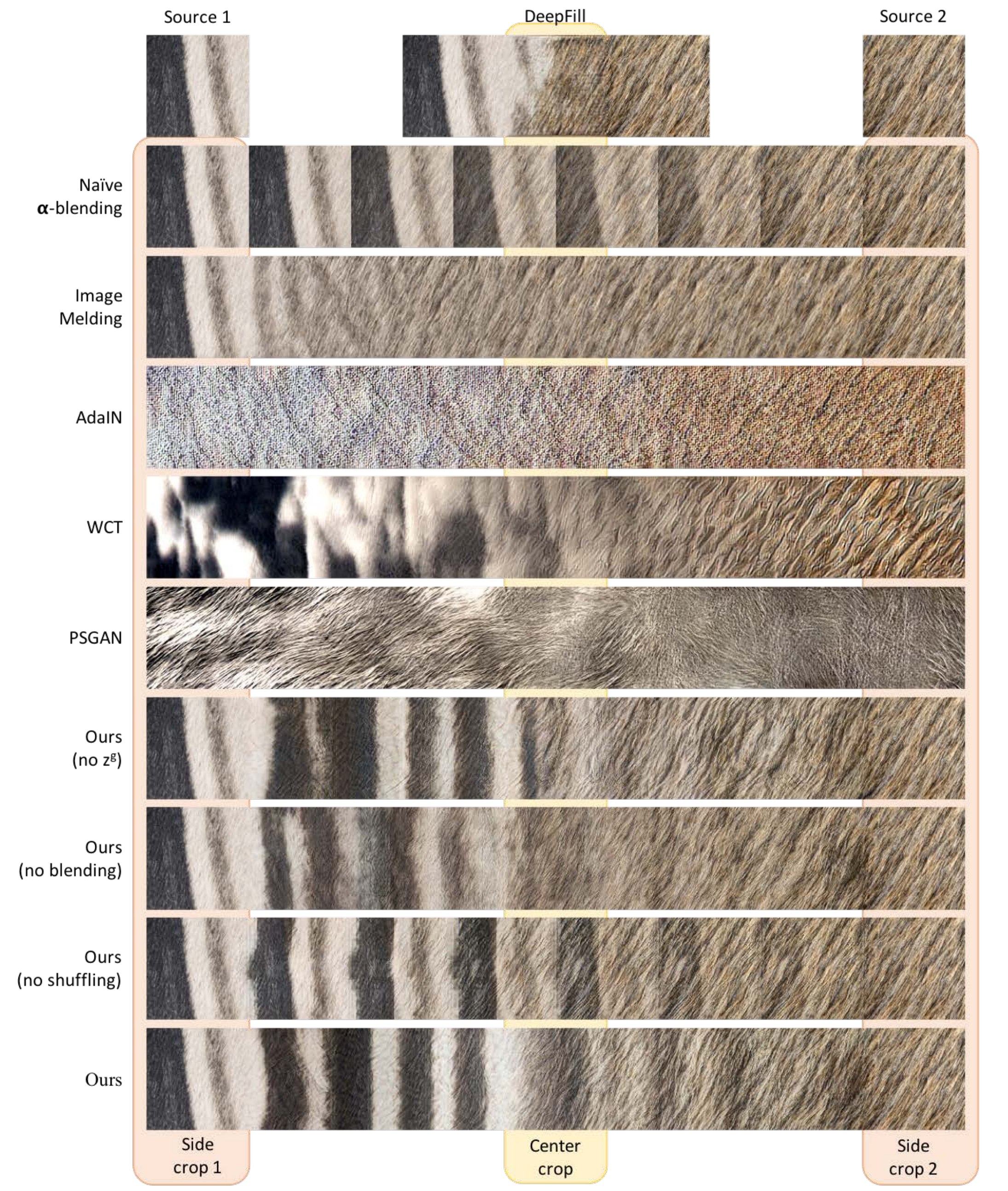}
\caption{Qualitative demonstrations and comparisons of horizontal interpolation in the size of $128 \times 1024$ on the \emph{animal texture} samples. We use the two side crops with the orange background for SPD measurement, and the center crop with the light yellow background for the other proposed quantitative evaluations.}
\label{fig:qual_eval_6}
\end{figure*}

\begin{figure*}
\centering
\includegraphics[width=1\linewidth]{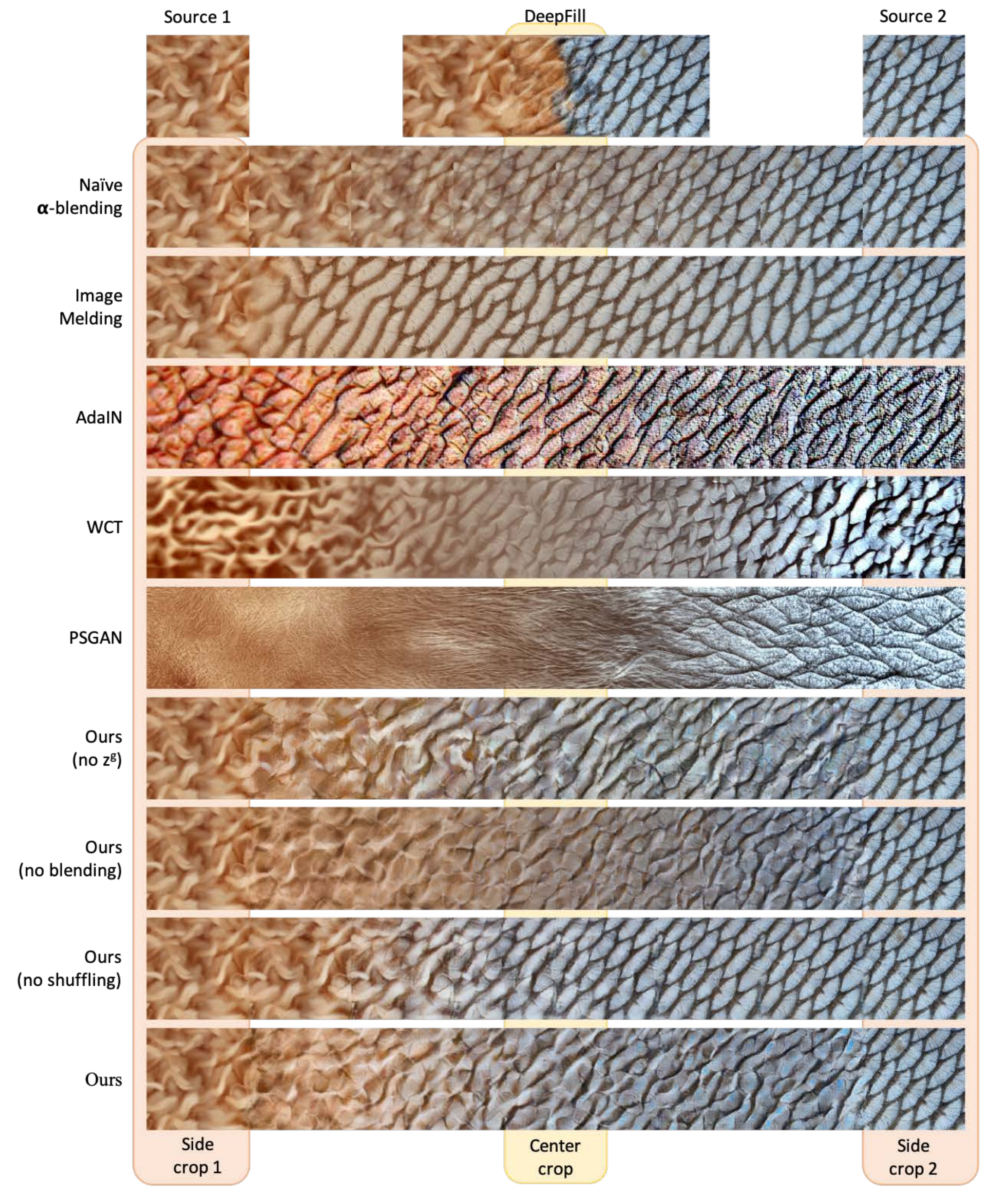}
\caption{A qualitative demonstration of one of our failure examples. When dealing with strong structural source texture on the right, our full method didn't outperform ours without random shuffling during training, and didn't outperform Image Melding either.}
\label{fig:qual_eval_failure}
\end{figure*}

\end{document}